\documentclass[letterpaper, 10 pt, journal, twoside]{IEEEtran}
\usepackage{times}
\usepackage[pdftex]{graphicx}
\usepackage{subfigure}
\usepackage{amsmath,amssymb,amsopn,amstext,amsfonts}
\usepackage{cancel}
\usepackage[space]{cite}
\usepackage{pdfsync}
\usepackage{balance}
\usepackage{color}
\usepackage{mathtools}
\usepackage{algpseudocode}
\usepackage[ruled,vlined,linesnumbered]{algorithm2e}

\algnewcommand\algorithmicforeach{\textbf{for each}}
\algdef{S}[FOR]{ForEach}[1]{\algorithmicforeach\ #1\ \algorithmicdo}

\usepackage{bm}
\usepackage{subfigure}
\usepackage{diagbox}
\usepackage{epstopdf}
\usepackage{pifont}
\usepackage{fixltx2e}
\usepackage{amsmath}
\usepackage{multirow}
\usepackage{url}
\usepackage{verbatim}
\usepackage{footmisc}
\usepackage{mathabx}	
\usepackage{physics}
\usepackage[linkcolor=black,citecolor=black,urlcolor=black,colorlinks=true]{hyperref}
\usepackage{subfigure}
\usepackage{color}
\usepackage{amssymb,mathrsfs,amsmath}
\usepackage[skip=3pt, font={small}]{caption} % example skip set to 2pt
\usepackage{tabularx}
\usepackage{tikz}
\usepackage{amsthm}
\usepackage{wrapfig}
\usepackage{tabulary}
\usepackage{booktabs}
\newtheoremstyle{mystyle}%                % Name
{}%                                     % Space above
{}%                                     % Space below
{\itshape}%                                     % Body font
{}%                                     % Indent amount
{\bfseries}%                            % Theorem head font
{.}%                                    % Punctuation after theorem head
{ }%                                    % Space after theorem head, ' ', or \newline
{\thmname{#1}\thmnumber{ #2}\thmnote{ (#3)}}%                                     % Theorem head spec (can be left empty, meaning `normal')

\theoremstyle{mystyle}

\graphicspath{{figures/}}
\DeclareGraphicsExtensions{.png,.jpg,.eps,.pdf}
\def\BibTeX{{\rm B\kern-.05em{\sc i\kern-.025em b}\kern-.08em
		T\kern-.1667em\lower.7ex\hbox{E}\kern-.125emX}}
\title{\Large R$^3$LIVE: A Robust, Real-time, RGB-colored, LiDAR-Inertial-Visual tightly-coupled state Estimation and mapping package
}

\author{
	Jiarong Lin and Fu Zhang % stops a space
	\thanks{J. Lin and F. Zhang are with the Department of Mechanical Engineering, The University of Hong Kong, Hong Kong SAR, China. {\tt\small $\{$jiarong.lin,  fuzhang$\}$@hku.hk}}
}%

\begin{document}
	\maketitle
	\begin{abstract}
		In this letter, we propose a novel LiDAR-Inertial-Visual sensor fusion framework termed R$^3$LIVE, which takes advantage of measurement of LiDAR, inertial, and visual sensors to achieve robust and accurate state estimation. R$^3$LIVE is contained of two subsystems, the LiDAR-inertial odometry (LIO) and visual-inertial odometry (VIO). The LIO subsystem (FAST-LIO) takes advantage of the measurement from LiDAR and inertial sensors and builds the geometry structure of (i.e. the position of 3D points) global maps. The VIO subsystem utilizes the data of visual-inertial sensors and renders the map's texture (i.e. the color of 3D points). More specifically, the VIO subsystem fuses the visual data directly and effectively by minimizing the frame-to-map photometric error. The developed system R$^3$LIVE is developed based on our previous work R$^2$LIVE, with careful architecture design and implementation. Experiment results show that the resultant system achieves more robustness and higher accuracy in state estimation than current counterparts (see our attached video\footnote{\url{https://youtu.be/j5fT8NE5fdg}}). 
		
		R$^3$LIVE is a versatile and well-engineered system toward various possible applications, which can not only serve as a SLAM system for real-time robotic applications, but can also reconstruct the dense, precise, RGB-colored 3D maps for applications like surveying and mapping. Moreover, to make R$^3$LIVE more extensible, we develop a series of offline utilities for reconstructing and texturing meshes, which further minimizes the gap between R$^3$LIVE and various of 3D applications such as simulators, video games and etc (see our demos video\footnote{\url{https://youtu.be/4rjrrLgL3nk}}). To share our findings and make contributions to the community, we open source R$^3$LIVE on our Github\footnote{\url{https://github.com/hku-mars/r3live}\label{foot_github}}, {including all of our codes, software utilities, and the mechanical design of our device}.
	\end{abstract}
	\vspace{-0.3cm}
%	\begin{IEEEkeywords}
%	SLAM, Mapping, Sensor Fusion.
%	\end{IEEEkeywords}

	%\note{In this One-Page abstract}, we present an overview of our open-source ros-package: the LOAM-Livox\footnote{\url{https://github.com/hku-mars/loam_livox}}, a robust LiDAR Odometry and mapping (LOAM) package for Livox LiDAR.
%	\vspace{-0.8cm}

\begin{figure}[t]
	\centering
	\centering
	\includegraphics[width=1.0\linewidth]{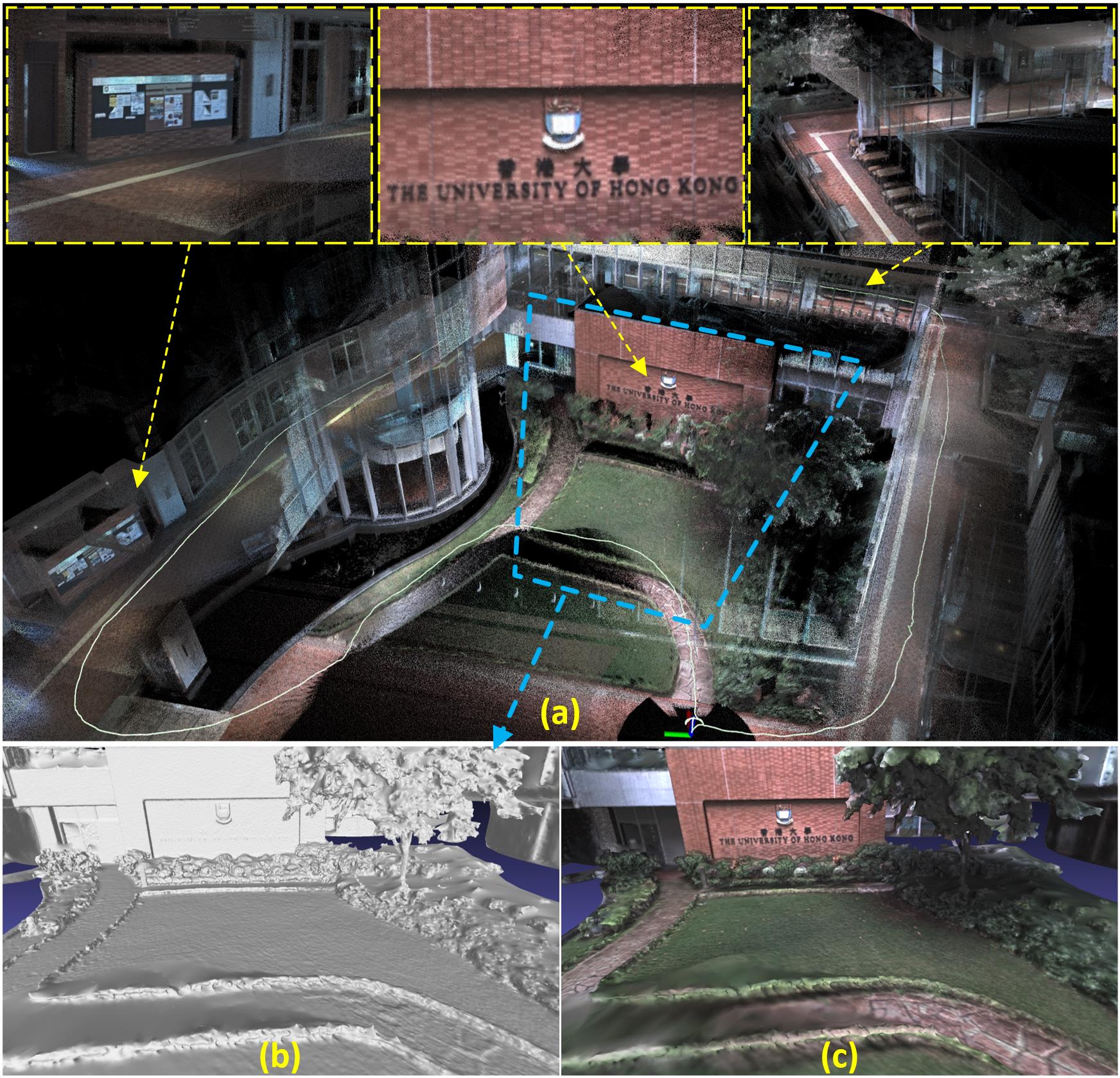}
	\caption[Caption for LOF]{(a) Our developed R$^3$LIVE is able to reconstruct a dense, 3D, RGB-colored point cloud of the traveled environment in real-time. The white path is our traveling trajectory for collecting the data. (b) offline reconstructed mesh, and the mesh after textured is rendered in (c).}
	\label{fig_cover}
	\vspace{-0.9cm}
\end{figure}
\section{Introduction}\label{sect_intro}
Recently, LiDAR sensors have been increasingly used in various robotic applications such as autonomous driving vehicles \cite{levinson2011towards}, drones \cite{bry2012state, gao2019flying, kong2021avoiding}, etc. Especially with the emergence of low-cost solid-state LiDARs (e.g., \cite{liu2020low}), more  applications \cite{xu2020fast,lin2020decentralized, liu2020balm, LiDARcalib, yuan2021pixel} based on these LiDARs propel the development of robotic community. However, for LiDAR-based SLAM systems, they would easily fail in those scenarios with no enough geometry features, especially for solid-state LiDARs which typically have limited FoV \cite{lin2020loam}. {To address this problem, fusing LiDAR with other sensors such as camera\cite{r2live, lvisam,zuo2019lic, zuo2020lic} and Ultra-wideband (UWB)\cite{zhen2019estimating, zhou2020uwb} can improve the system's robustness and accuracy. In particular, various of LiDAR-Visual fusion frameworks have been proposed recently in the robotics community\cite{debeunne2020review}.}

{V-LOAM \cite{zhang2018laser} proposed by Zhang and Singh is one of the early works of LiDAR-Inertial-Visual systems, which leverage a loosely-coupled Visual-Inertial odometry (VIO) as the motion model for initializing the LiDAR mapping subsystem. Similarly, in \cite{shao2019stereo}, the authors propose a stereo visual-inertial LiDAR SLAM which incorporates the tightly-coupled stereo visual-inertial odometry with the LiDAR mapping and LiDAR enhanced visual loop closure. More recently, Wang {\it et al} propose DV-LOAM \cite{wang2021dv}, which is a direct Visual-LiDAR fusion framework. The system first utilizes a two-staged direct visual odometry module for efficient coarse state estimate, then the coarse pose is refined with the LiDAR mapping module, and finally the loop-closure module is leveraged for correcting the accumulated drift. The above systems fuse LiDAR-inertial-visual sensors at a level of loosely-coupled, in which the LiDAR measurement are not jointly optimized together with the visual or inertial measurements.
}

{
	More tightly-coupled LiDAR-Inertial-Visual fusion frameworks have been proposed recently. For example, LIC-fusion\cite{zuo2019lic} proposed by Zuo \textit{et al} is a tightly-coupled LiDAR-Inertial-Visual fusion framework that combines IMU measurements, sparse visual features, LiDAR features with online spatial and temporal calibration within the Multi-State Constrained Kalman Filter (MSCKF) framework. To further enhance the robustness of the LiDAR scan matching, their subsequent work termed LIC-Fusion 2.0\cite{zuo2020lic} proposes a plane-feature tracking algorithm across multiple LiDAR scans within a sliding-window and refines the pose trajectories within the window. Shan \textit{et al} in \cite{lvisam} propose LVI-SAM fuses LiDAR-Visual-Inertial senor via a tightly-coupled smooth and mapping framework, which is built atop a factor graph. The LiDAR-Inertial and Visual-Inertial subsystems of LVI-SAM can function independently when failure is detected in one of them, or jointly when enough features are detected. Our previous work R$^2$LIVE \cite{r2live} tightly fuses the data of LiDAR-Inertial-Visual sensors, extract LiDAR and sparse visual features, estimates the state by minimizing the feature re-projection error within the framework of error-state iterated Kalman-filter for achieving the real-time performance, meanwhile improves the overall visual mapping accuracy with a sliding window optimization. R$^2$LIVE is able to run in various challenging scenarios with aggressive motions, sensor failures, and even in narrow tunnel-like environments with a large number of moving objects and with a small LiDAR FoV. 
}

In this paper, we attack the problem of real-time simultaneous localization, 3D mapping, and map renderization based on tightly-coupled fusion of LiDAR, inertial, and visual measurements. Our contributions are:

\begin{itemize}
	\item We propose a real-time simultaneous localization, mapping, and colorization framework. The presented framework consists of a LiDAR-inertial odometry (LIO) for reconstructing geometry structure and a visual-inertial odometry (VIO) for texture rendering. The overall system is able to reconstruct a dense, 3D, RGB-colored point cloud of the environment in real-time (Fig. \ref{fig_cover} (a))
	\item We propose a novel VIO system based on the RGB-colored point cloud map. The VIO estimates the current state by minimizing the photometric error between the RGB color of an observed map point and its measured color in the current image. Such a process requires no salient visual feature in the environment and saves the corresponding processing time (e.g. features detection and extraction), which makes our proposed system more robust especially in texture-less environments.
	\item We implement the proposed methods into a complete system, R$^3$LIVE, which is able to construct the dense, precise, 3D, RGB-colored point cloud maps of the environment in real-time and with low-drift. The overall system is validated in various indoor and outdoor environments. Results show that our system drifts only $0.16$ meters in translation and $3.9$ degree in rotation after traveling up to 1.5 kilometers. 
	\item We open source our system on our Github. We also develop several offline tools for reconstructing and texturing meshes from colorized point cloud (see Fig. \ref{fig_cover} (b) and (c)). These software utilities and mechanical design of our devices are also made open source to benefit the possible applications.
\end{itemize}

\begin{figure}
	\centering
	\centering
	\includegraphics[width=1.05\linewidth]{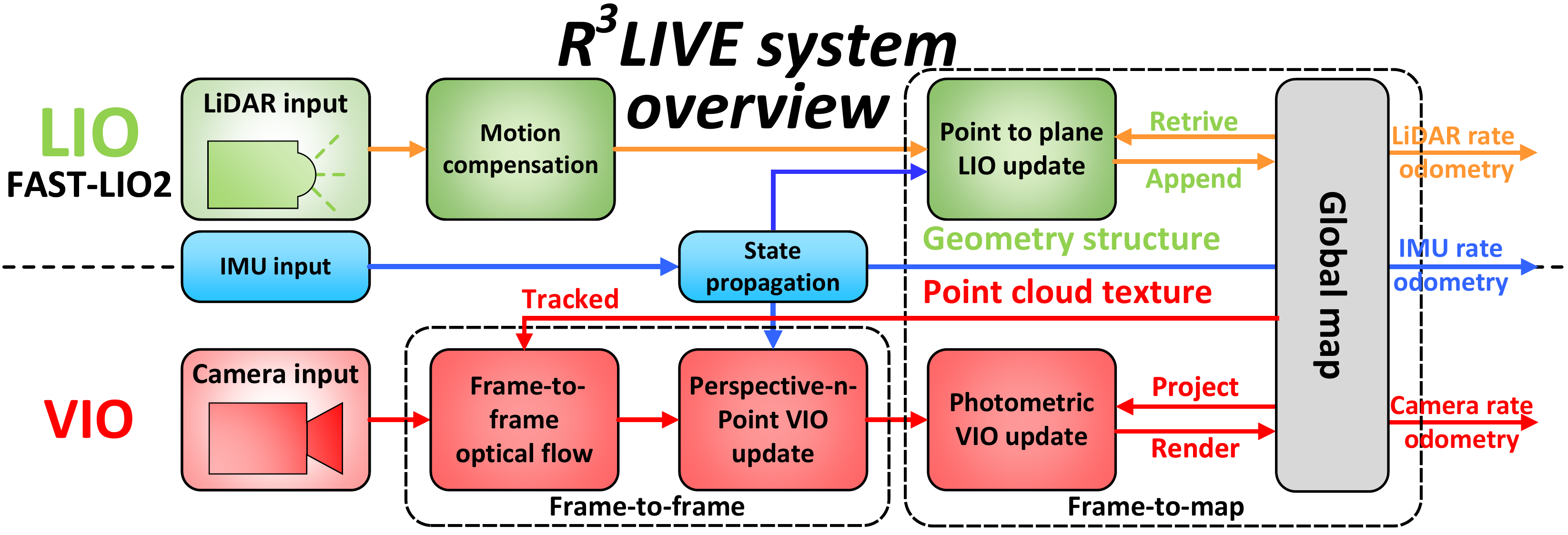}
\caption{The overview of our proposed system.}
	\label{fig_overview}
\end{figure}
	%	\vspace{-1.0cm}
%\end{figure*}
\section{The system overview}
The overview of our system is shown in Fig. \ref{fig_overview}, our proposed framework contains two subsystems: the LIO subsystem (upper part) and the VIO subsystem (the lower part). The LIO subsystem constructs the geometry structure of a global map, which registers the input LiDAR scan, and estimates the system's state by minimizing the point-to-plane residuals. The VIO subsystem builds the map's texture, which renders the RGB color of each point with the input image, updates the system state by minimizing the frame-to-frame PnP reprojection error and the frame-to-map photometric error. 

\section{Notation}
Throughout this paper, we use notations shown in TABLE. \ref{table_I_NOMENCLATURE}, which have been introduced in the previous work R$^2$LIVE\cite{r2live}.
	\begin{table}[htbp]
		\caption{NOMENCLATURE}
%		\scriptsize
%		\footnotesize
		\setlength\tabcolsep{2.5pt}
		\begin{tabular}{r l p{10cm} }
			\toprule
			\textbf{Notation} & \textbf{\hspace{2cm} Explanation} \\
			\toprule
			&\hspace{2cm}{\textit{Expression}}\\
			\toprule
			$\boxplus / \boxminus$ &  The encapsulated ``boxplus'' and \\  & \quad``boxminus'' operations on manifold  \\		
			${^G}(\cdot)$ &  				The coordinate of vector $(\cdot)$ in global frame \\		
			${^C}(\cdot)$ &  				The coordinate of vector $(\cdot)$ in camera frame \\		
			$\mathtt{Exp}(\cdot) / \mathtt{Log}(\cdot)$ & The Rodrigues' transformation between the\\ & \quad rotation matrix and rotation vector \\
			$\delta\left(\cdot\right)$ & The estimated error of $(\cdot)$. It is the minimum \\& \quad  parameterization that characterized in the tangent space.\\
			$\boldsymbol{\Sigma}_{(\cdot)}$ & The covariance matrix of vector $(\cdot)$ \\
			\toprule
			&\hspace{2cm}{\textit{Variable}} \\
			\toprule
			%			$i,j,k,l,n$    & Commonly used integers.  \\
			%			$\mathbf{x}$ & The full state vector \\
			%			$\hat{\mathbf{x}}$ & The estimated state\\
			%			$\delta{\hat{\mathbf{x}}}$ & The state esimation error of $\hat{\mathbf{x}}$ \\
			$\mathbf{b}_{\mathbf{g}}, \mathbf{b}_{\mathbf{a}}$ &  The bias of gyroscope and accelerometer \\
			${^G}\mathbf{g}$ & The gravitational acceleration in global frame \\
			${^G}\mathbf{v}$ & The linear velocity in global frame \\
			$({^G}\mathbf{R}_{I}, {^G}\mathbf{p}_{I})$ & The attitude and position of the IMU w.r.t. global frame \\
			$({^I}\mathbf{R}_{C}, {^I}\mathbf{p}_{C})$ & The extrinsic value between camera and IMU \\
			$\mathbf{x}$ &  The full state vector \\
			$\hat{\mathbf{x}}$ & The prior estimation of $\mathbf{x}$ \\
			$\check{\mathbf{x}}$ & The current estimate of $\mathbf{x}$ in each ESIKF iteration  \\  
			\bottomrule \\
			\label{table_I_NOMENCLATURE}
		\end{tabular}
		\vspace{-1.1cm}
	\end{table}

\subsection{State vector}
In our work, we define the full state vector $\mathbf{x}\in \mathbb{R}^{29}$ as: 
\begin{small}
	\begin{align}
		\hspace{-0.0cm}\mathbf{x} = 
		\begin{bmatrix}
			^G\mathbf{R}_{I}^T, {^G}\mathbf{p}_{I}^T,  {^G}\mathbf{v}^T, \mathbf{b}_{\mathbf{g}}^T, \mathbf{b}_{\mathbf{a}}^T, {^G}\mathbf{g}^{T}, {^I}\mathbf{R}_{C}^T, {^I}\mathbf{p}_{C}^T, {^I}t_{C} , \boldsymbol{\phi}^T
		\end{bmatrix}^T 
	\end{align}
\end{small}
\hspace{-0.07cm}where ${^G}\mathbf{g}\in \mathbb{R}^{3}$ is the gravity vector expressed in global frame (i.e. the first LiDAR frame), ${^I}t_{C}$ is the time-offset between IMU and camera while LiDAR is assumed to be synced with IMU already, $\boldsymbol{\phi} = \begin{bmatrix}
	f_x, f_y, c_x, c_y		
\end{bmatrix}^T$ is the camera intrinsic vector, with $(f_x, f_y)$ the pixel focal length and $(c_x, c_y)$ the offsets of the principal point from the top-left corner of the image plane.

\subsection{Maps representation}
Our map is made up of voxels and points, where points are contained in voxels and are the minimum elements of maps.

\subsubsection{Voxels}
For fast finding the points in maps for rendering and tracking (see Section. \ref{sect_render_texture_of_maps} and Section. \ref{sect_update_tracking_pts}) in our VIO subsystem, we design a fix-size (e.g. $0.1m\times 0.1m \times 0.1m $) container named voxel. If a voxel has points appended recently (e.g. in recent 1 second), we mark this voxel as \textit{activated}. Otherwise, this voxel is marked as \textit{deactivated}.

\subsubsection{Points}

In our work, a points $\mathbf{P}$ is a vector with size 6:
\begin{align}
%	\mathbf{P} = \left[ \mathbf{P}_x, \mathbf{P}_y, \mathbf{P}_z, \mathbf{P}_r, \mathbf{P}_g, \mathbf{P}_b \right]^T = \left[{^G}\mathbf{P}^T, {^C}\mathbf{P}^T \right]^T
	\mathbf{P} = \left[ {^G}\mathbf{p}_x, {^G}\mathbf{p}_y, {^G}\mathbf{p}_z, \mathbf{c}_r, \mathbf{c}_g, \mathbf{c}_b \right]^T = \left[{^G}\mathbf{p}^T, \mathbf{c}^T \right]^T
\end{align}
where the first $3 \times 1$ sub-vector ${^G}\mathbf{p} = \left[ \mathbf{P}_x, \mathbf{P}_y, \mathbf{P}_z\right]^T $ denotes the point 3D position in the global frame, and the second  $3\times 1$ vector $\mathbf{c} = \left[ \mathbf{c}_r, \mathbf{c}_g, \mathbf{c}_b\right]^T $ is the point RGB color. Besides, we also record other informations of this point, such as the $3\times 3$ covariance matrix $\boldsymbol{\Sigma}_{\mathbf{p}}$ and $\boldsymbol{\Sigma}_{\mathbf{c}}$, which denote the estimated errors of ${^G}\mathbf{p}$ and $\mathbf{c}$, respectively, and the timestamps when this point have been created and rendered.

\section{LiDAR-Inertial odometry subsystem}\label{sect_LIO_subsystem}
As shown in Fig. \ref{fig_overview}, the LIO subsystem of R$^3$LIVE constructs the geometry structure of the global map. For an incoming LiDAR scan, the motion distortion due to the continuous movement within the frame is compensated by IMU backward propagation, as shown in \cite{xu2020fast}. Then, we leverage the error-state iterated Kalman filter (ESIKF) minimizing the point-to-plane residuals to estimate the system's state. Finally, with the converged state, the points of this scan are appended to the global map and mark the corresponding voxels as {\it activated} or {\it deactivated}. The accumulated 3D points in the global map form the geometry structure, which is also used for providing the depth for our VIO subsystem. For the detailed implementations of the LIO subsystem in R$^3$LIVE, we refer our readers to our previous related works \cite{fastlio2, r2live}.

\section{Visual-Inertial odometry subsystem}
Our VIO subsystem renders the texture of a global map, with estimating the system state by minimizing the photometric error. To be more specific, we project a certain number of points (i.e. tracked points) from the global map to the current image, then iteratively estimate the system state within the framework of ESIKF by minimizing the photometric error of these points. The tracked map points are sparse for the sake of efficiency, which usually requires to build a pyramid\footnote{\url{https://en.wikipedia.org/wiki/Pyramid_(image_processing)}} of the input image. The pyramid is however not invariant to either translation or rotation that needs also to be estimated. In our proposed framework, we utilize the color of a single map point to compute the photometric error. The color, which is rendered simultaneously in the VIO, is an inherent property of the map point and is invariant to both translation and rotation of the camera. To ensure a robust and fast convergence, we design a two-step pipeline shown in Fig. \ref{fig_overview}, where we firstly leverage a frame-to-frame optical flow to track map points and optimize the system's state by minimizing the Perspective-n-Point (PnP) projection error of the tracked map points (Section \ref{section_frame_to_frame_vio}). Then, we further refine the system's state estimation by minimizing the frame-to-map photometric error among the tracked points (Section \ref{Frame_to_map}). With the converged state estimate and the raw input image, we perform the texture rendering to update the colors of points in the global map (Section \ref{sect_render_texture_of_maps}).

\subsection{Frame-to-frame Visual-Inertial odometry}\label{section_frame_to_frame_vio}

Assume we have tracked $m$ map points $\boldsymbol{\mathcal{P}} = \{ \mathbf{P}_1, ..., \mathbf{P}_m \}$ in last image frame $\mathbf{I}_{k-1}$, and their projection on the $\mathbf{I}_{k-1}$ are $\{ \boldsymbol{\rho}_{1_{k-1}},...,  \boldsymbol{\rho}_{m_{k-1}} \}$, we leverage the Lucas$-$Kanade optical flow to find out their location in the current image frame $\mathbf{I}_k$, denoted as $\{ \boldsymbol{\rho}_{1_k},...,  \boldsymbol{\rho}_{m_k}\}$. Then, we calculate and optimize their projection error to estimate the state via an ESIKF. 

\subsubsection{Perspective-n-Point projection error}\label{section_PnP_projection_error}

\begin{figure}[t]
	\centering
	\includegraphics[width=0.7\linewidth]{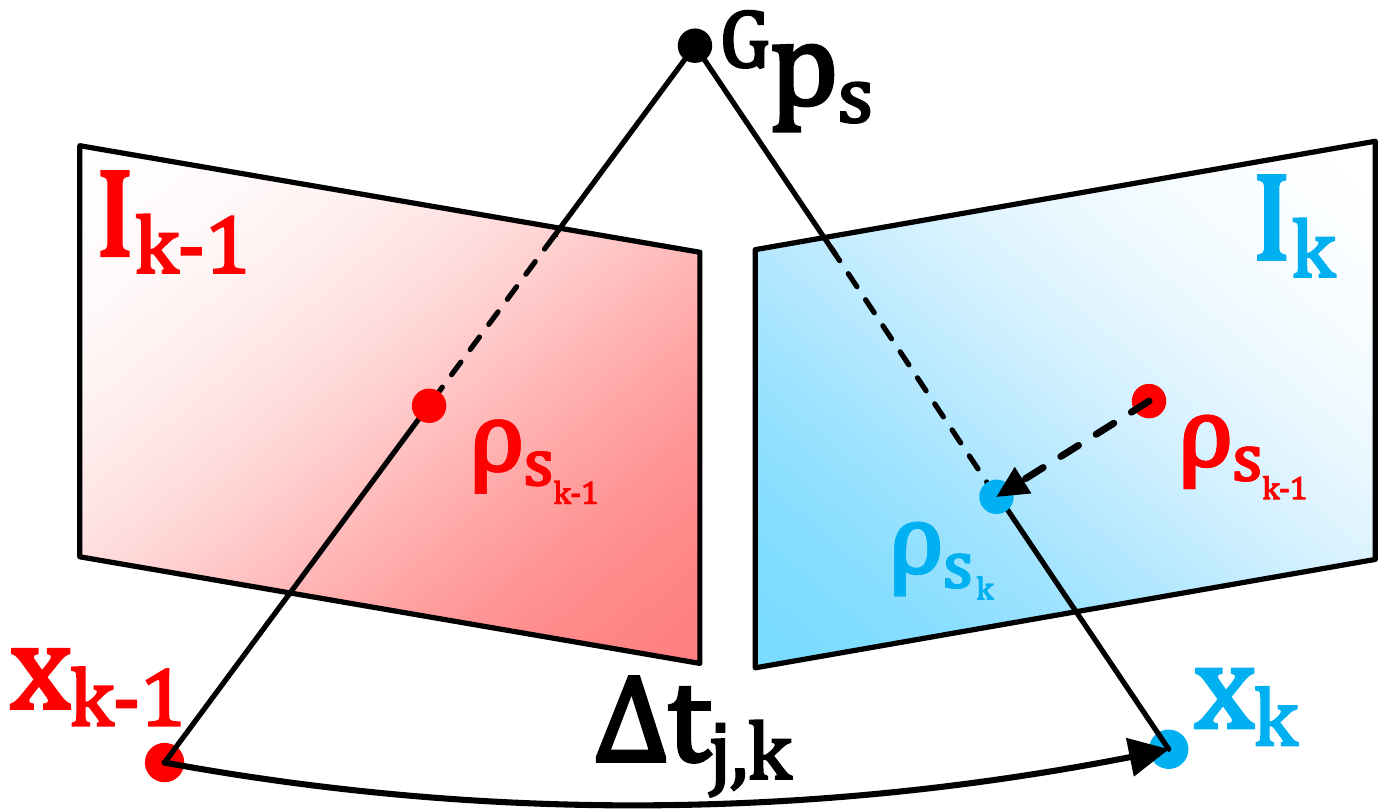}
	\caption{Our frame-to-frame ESIKF VIO update the system's state with minimizing the PnP projection error.}
	\label{fig_frame_to_frame}
	\vspace{-0.8cm}
\end{figure}

As shown in Fig. \ref{fig_frame_to_frame}, we take the $s$-th point $\mathbf{P}_s =  \left[ {^G}\mathbf{p}_s^T, \mathbf{c}_s^T \right]^T  \in \boldsymbol{\mathcal{P}}$ as example, the projection error is calculated as follow $\mathbf{r}\left(\check{\mathbf{x}}_{k}, \boldsymbol{\rho}_{s}^k , {^G\mathbf{p}_{s}}\right)$:
%\begin{small}
\begin{align}
%	\mathbf{P}_s =  \left[ {^G}\mathbf{p}_s^T, \mathbf{c}_s^T \right]^T  	=\left[ {^G}\mathbf{p}_{x_s}, {^G}\mathbf{p}_{y_s}, {^G}\mathbf{p}_{z_s}, \mathbf{c}_{r_s}, \mathbf{c}_{g_s}, \mathbf{c}_{b_s} \right]^T \\
	\begin{split}
	  {^C} \mathbf{p}_s  &= \left[ {^C}\mathbf{p}_{x_s}, {^C}\mathbf{p}_{y_s}, {^C}\mathbf{p}_{z_s} \right]^T = 	\left({^G\check{\mathbf{R}}_{I_{k}}} \cdot {^I\check{\mathbf{R}}_{C_{k}}} \right)^T  \cdot {{^G}{\mathbf{p}}_s}  \\
	& \quad \quad \quad -{^I\check{\mathbf{R}}_{C_{k}}^T} \cdot {^I\check{\mathbf{p}}_{C_{k}}} - \left({^G\check{\mathbf{R}}_{I_{k}}} \cdot {^I\check{\mathbf{R}}_{C_{k}}}\right)^T  \cdot {{^G}\check{\mathbf{p}}_{I_{k}}}  \label{eq_hat_C_p_s}
	\end{split} \\
	\mathbf{r}&\left(\check{\mathbf{x}}_{k}, \boldsymbol{\rho}_{s_k} , {^G\mathbf{p}_{s}}\right) = 
	\boldsymbol{\rho}_{s_k} - \boldsymbol{\pi}({^C\mathbf{p}_{s}}, \check{\mathbf{x}}_k ) \label{eq_frame_to_map_projection_err} 
%	{^G}\mathbf{p}_s &= 
\end{align}
%\end{small}
where $\check{\mathbf{x}}_{k}$ is the current state estimate of $\mathbf{x}$ in each ESIKF iteration, $ \boldsymbol{\pi}({^C\mathbf{p}_{s}}, \check{\mathbf{x}}_k ) \in \mathbb{R}^2 $ is computed as follow:
\begin{align}
\begin{split}
	\boldsymbol{\pi}({^C\mathbf{p}_{s}}, \check{\mathbf{x}}_k )  &= \left[ \check{f}_{x_k} \dfrac{{^C}\mathbf{p}_{x_s}}{{^C}\mathbf{p}_{z_s}} + \check{c}_{x_k} ,~  \check{f}_{y_k} \dfrac{{^C}\mathbf{p}_{y_s}}{{^C}\mathbf{p}_{z_s}} + \check{c}_{y_k}\right]^T \\ 
&	+ \dfrac{{^I}\check{t}_{C_k}}{\Delta t_{k-1,k}}( \boldsymbol{\rho}_{s_k} - \boldsymbol{\rho}_{s_{k-1}} ) \label{eq_projection}
\end{split}
\end{align}
where  $\Delta t_{k-1,k}$ is the time internal between last image frame $\mathbf{I}_{k-1}$ and current image frame $\mathbf{I}_k$. Notice that in (\ref{eq_projection}), the first item is the pin-hole projection function and the second one is the online-temporal correction factor \cite{qin2018online}.

The measurement noise in the residual (\ref{eq_frame_to_map_projection_err}) consists of two sources: one is the pixel tracking error in $\boldsymbol{\rho}_{s_k}$ and the other lies in the map point location error ${^G\mathbf{p}_{s}}$, 
\begin{align}
	{^G\mathbf{p}_{s}}	= {^G\mathbf{p}_{s}^{\mathtt{gt}}}  + \mathbf{n}_{\mathbf{p}_{s}},~& \mathbf{n}_{\mathbf{p}_{s}} \sim \mathcal{N}(\mathbf{0}, \boldsymbol{\Sigma}_{\mathbf{n}_{\mathbf{p}_{s}}})  \\
	\boldsymbol{\rho}_{s_k} = {\boldsymbol{\rho}_{s_k}^{\mathtt{gt}}} + \mathbf{n}_{\boldsymbol{\rho}_{s_k}},~& \mathbf{n}_{\boldsymbol{\rho}_{s_k}} \sim \mathcal{N}(\mathbf{0}, \boldsymbol{\Sigma}_{\mathbf{n}_{\boldsymbol{\rho}_{s_k}}}) 
\end{align}
where $ {^G\mathbf{p}_{s}^{\mathtt{gt}}}$ and ${\boldsymbol{\rho}_{s_k}^{\mathtt{gt}}}$ are the true values of ${^G\mathbf{p}_{s}}$ and $\boldsymbol{\rho}_{s_k}$, respectively. Then, we obtain the first order Taylor expansion of the true zero residual $\mathbf{r}(\mathbf{x}_{k}, \boldsymbol{\rho}_{s_k}^{\mathtt{gt}}, {^G\mathbf{p}_{s}^{\mathtt{gt}}})$ as:
\begin{equation}
	\begin{split}
	\hspace{-0.3cm}	\mathbf 0 &= \mathbf{r}(\mathbf{x}_{k}, \boldsymbol{\rho}_{s_k}^{\mathtt{gt}}, {^G\mathbf{p}_{s}^{\mathtt{gt}}}) \approx \mathbf{r}\left(\check{\mathbf{x}}_{k}, \boldsymbol{\rho}_{s_k} , {^G\mathbf{p}_{s}}\right) + \mathbf{H}^{r}_{s} \delta \check{\mathbf{x}}_{k} + \boldsymbol{\alpha}_{s} \label{eq_visual_meas}
	\end{split}
\end{equation}
where $\boldsymbol{\alpha}_{s} \sim \mathcal{N}(\mathbf{0}, {\boldsymbol{\Sigma}_{\boldsymbol{\alpha}_{s}}}  )$ and
%\begin{footnotesize}
\begin{align}
	\hspace{-0.2cm}\mathbf{H}^{r}_{s}  &=\left.\dfrac{ \partial  \mathbf{r}_c(\check{\mathbf{x}}_{k} \boxplus \delta{\check{\mathbf{x}}}_{k},  \boldsymbol{\rho}_{s_k} , {^G\mathbf{p}_{s}} ) }{\partial  \delta{\check{\mathbf{x}}}_{k}} \right|_{\delta{\check{\mathbf{x}}}_{k} = \mathbf{0}}  \\
	{\boldsymbol{\Sigma}_{\boldsymbol{\alpha}_{s}}} &= \boldsymbol{\Sigma}_{\mathbf{n}_{\boldsymbol{\rho}_{s_k}}} + {\mathbf{F}^r_{{{\mathbf{p}}}_{s}}} \boldsymbol{\Sigma}_{\mathbf{p}_{s}} {\mathbf{F}^r_{{{\mathbf{p}}}_{s}}}^T \label{eq_def_Hc} \\
	\mathbf{F}^r_{{{\mathbf{p}}}_{s}} &=   
	\dfrac{ \partial  \mathbf{r}(\check{\mathbf{x}}_{k}, \boldsymbol{\rho}_{s_k} , {^G\mathbf{p}_{s}})  }{\partial  {^{G}{\mathbf{p}}}_{s} }.
\end{align}

\subsubsection{Frame-to-frame VIO ESIKF update}\label{sec:update}
%\subsection{Update of error-state iterated Kalman filter}\label{sec:update}

Equation (\ref{eq_visual_meas}) constitutes an observation distribution for ${\mathbf{x}}_{k}$ (or equivalently $\delta \check{\mathbf{x}}_{k} \triangleq {\mathbf{x}}_{k} \boxminus \check{\mathbf{x}}_{k}$), which can be combined with the prior distribution from the IMU propagation to obtain the maximum a posteriori (MAP) estimation of $\delta \check{\mathbf{x}}_{k}$:
%\begin{small}
	\begin{align}
	\begin{split}
		%       \begin{split}
		\hspace{-0.5cm}\mathop{\min}_{\delta \check{\mathbf{x}}_{k}} &\left(  \left\| \check{\mathbf{x}}_{k} \boxminus \hat{\mathbf{x}}_{k}  + \boldsymbol{\mathcal{H}} \delta{\check{\mathbf{x}}}_{k}   \right\|_{\boldsymbol{\Sigma}_{\delta \hat{\mathbf{x}}_{k}} }^2 \right.  \\
		+&\left. \sum\nolimits_{s=1}^{m}\left\|  {  \mathbf{r}\left(\check{\mathbf{x}}_{k}, \boldsymbol{\rho}_{s_k} , {^G\mathbf{p}_{s}}\right) + \mathbf{H}^{r}_{s} \delta \check{\mathbf{x}}_{k} } \right\|^2_{{\boldsymbol{\Sigma}_{\boldsymbol{\alpha}_{s}}}}  \right) 
		%       \end{split}
		\label{eq_optimial_frame_to_frame}
	\end{split}
	\end{align}
%\end{small}
where $\left\| \mathbf{x} \right\|_{\boldsymbol{\Sigma}}^2 = \mathbf{x}^T \boldsymbol{\Sigma}^{-1} \mathbf{x}  $ is the squared Mahalanobis distance with covariance $\boldsymbol{\Sigma}$,  $\hat{\mathbf{x}}_{k}$ is the IMU propagated state estimate, $\boldsymbol{\Sigma}_{\delta \hat{\mathbf{x}}_{k}}$ is the IMU propagated state covariance, and $\boldsymbol{\mathcal{H}}$ is the Jacobian matrix when projecting the state error from the tangent space of $\hat{\mathbf{x}}_{k}$ to the tangent space of $\check{\mathbf{x}}_{k}$. {The detailed derivation of first item in (\ref{eq_optimial_frame_to_frame}) can be found in Section. E of R$^2$LIVE\cite{r2live}}.

Denote:
%\begin{small}
	\begin{align}
		\mathbf{H} &= 
		\begin{bmatrix}
			{\mathbf{H}^{r}_{1}}^T ,\dots, {\mathbf{H}^{r}_{m}}^T 
		\end{bmatrix}^T \label{eq_H_mat_frame_to_frame} \\
		\mathbf{R} &= 
		\text{diag}(
		\begin{matrix}
			\boldsymbol{\Sigma}_{\boldsymbol{\alpha}_{1}} , \dots ,\boldsymbol{\Sigma}_{\boldsymbol{\alpha}_{m}}
		\end{matrix}
		) \label{eq_R_mat_frame_to_frame} \\
		\check{\mathbf{z}}_{k} &= 
		\left[
		\mathbf{r}\left(\check{\mathbf{x}}_{k}, \boldsymbol{\rho}_{1_k} , {^G\mathbf{p}_{1}}\right)\dots,  
		\mathbf{r}\left(\check{\mathbf{x}}_{k}, \boldsymbol{\rho}_{m_k} , {^G\mathbf{p}_{m}}\right) \right] ^T \label{eq_z_mat_frame_to_frame} \\
		\mathbf{P} &= 
		\left(\boldsymbol{\mathcal{H}}\right)^{-1} \boldsymbol{\Sigma}_{\delta \hat{\mathbf{x}}_{k}} {\left(\boldsymbol{\mathcal{H}}\right)}^{-T} \label{eq_P_mat_frame_to_frame} 
	\end{align}
%\end{small}

Following \cite{xu2020fast}, we have the Kalman gain computed as:
\begin{align}
	\mathbf{K} = \left( \mathbf{H}^T\mathbf{R}^{-1}\mathbf{H} + \mathbf{P}^{-1} \right)^{-1}\mathbf{H}^T\mathbf{R}^{-1} \label{eq_K_mat_frame_to_frame}
\end{align}

Then we can update the state estimate as:
\begin{small}
	\begin{align}
		\check{\mathbf{x}}_{k} =& \check{\mathbf{x}}_{k}\boxplus \left( -\mathbf{K}\check{\mathbf{z}}_{k} - \left( \mathbf{I}- \mathbf{KH} \right) \left( \boldsymbol{\mathcal{H}} \right)^{-1} \left( \check{\mathbf{x}}_{k} \boxminus \hat{\mathbf{x}}_{k} \right) \right) \label{eq_x_k_update_frame_to_frame}
	\end{align}
\end{small}
%	The updated estimation is used to compute the residual of the next iteration, we repeat the process until state is covergence (i.e., $\mathbf{x}^{{i}+1}_{k}\boxminus \mathbf{x}_k^{i} < \epsilon$)  or reach our maximum allowance step.
The above process (Section \ref{section_PnP_projection_error} to Section \ref{sec:update}) is iterated until convergence (i.e., the update is smaller than a given threshold). Notice that such an iterated Kalman Filter is known to be equivalent to a Gauss-Newton optimization \cite{bell1993iterated}.

\begin{figure}[t]
	\centering
	\includegraphics[width=1.0\linewidth]{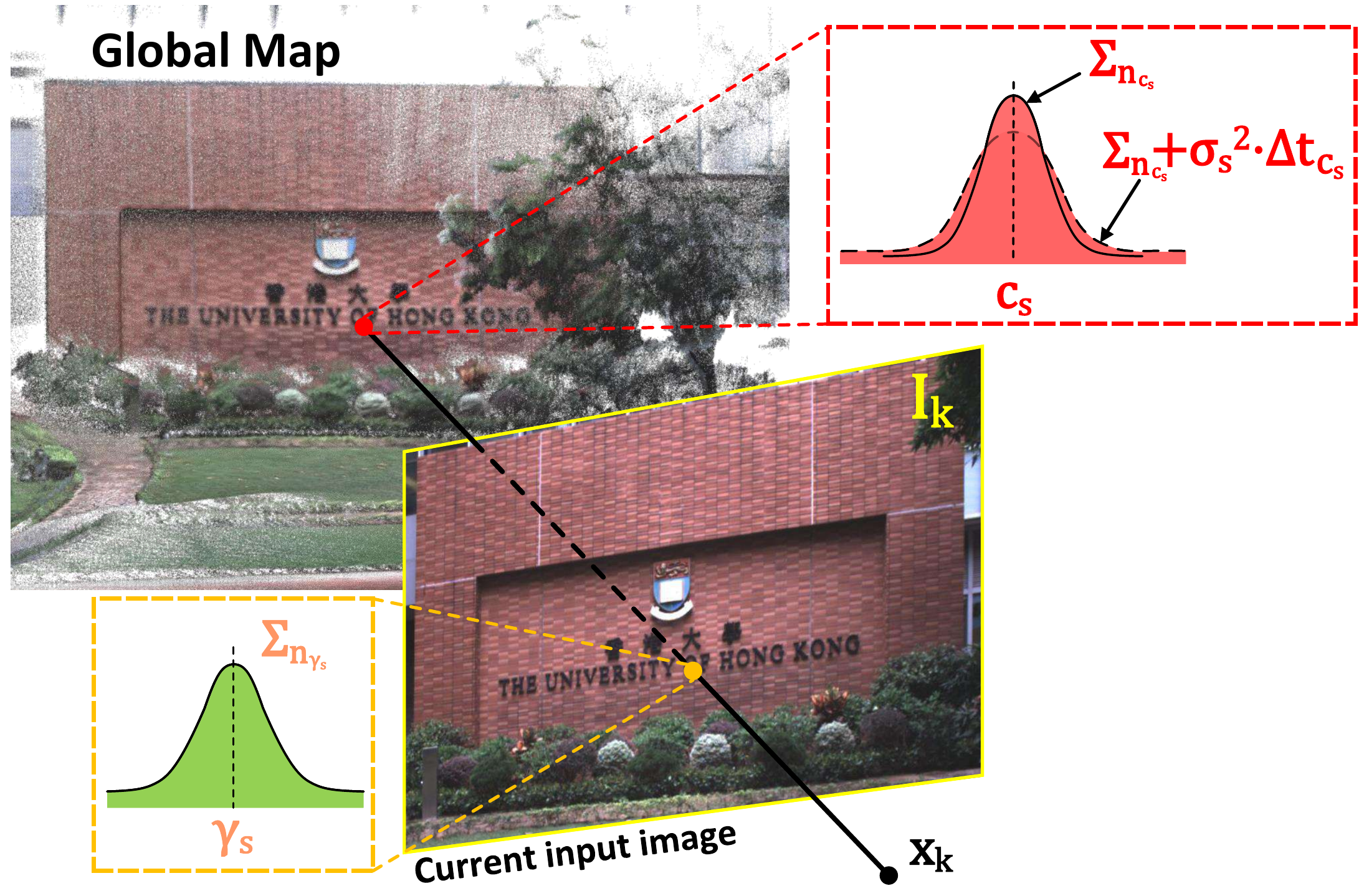}
	\caption{Our frame-to-map ESIKF VIO update the system state by minimizing the photometric error between the map point color and its measured color in the curren timage.}
	\label{fig_frame_to_frame_PE}
\end{figure}

\subsection{Frame-to-map Visual-Inertial odometry}\label{Frame_to_map}
\subsubsection{Frame-to-map photometric update}\label{sect_frame_to_map_PE}
After the frame-to-frame VIO update, we have a good estimated state $\check{\mathbf{x}}_k$, then we perform the frame-to-map VIO update by minimizing the photometric error of the tracked points to lower the drift.{As shown in Fig. \ref{fig_frame_to_frame_PE}, take the $s$-th tracked point $\mathbf{P}_s \in \boldsymbol{\mathcal{P}}$ as example, its frame-to-map photometric error $\mathbf{o}( \check{\mathbf{x}}_k, {^G}\mathbf{p}_s, \mathbf{c}_s  )$  is calculated as follow:
 \begin{align}
 	\mathbf{o}( \check{\mathbf{x}}_k, {^G}\mathbf{p}_s, \mathbf{c}_s  ) = \mathbf{c}_s  - \boldsymbol{\gamma}_s \label{eq_frame_to_map_photo_err}
 \end{align}
where $\mathbf{c}_s$ is the point color saved in the global map, $\boldsymbol{\gamma}_s$ is the color observed in the current image $\mathbf{I}_{k}$. To obtain the observed color $\boldsymbol{\gamma}_s$ and its covariance $\boldsymbol{\Sigma}_{\mathbf{n}_{{\boldsymbol{\gamma}_s}}}$, we predict the point location  $\tilde{\boldsymbol{\rho}}_{s_k} = \boldsymbol{\pi}({^C\mathbf{p}_{s}}, \check{\mathbf{x}}_k )$ on the current image $\mathbf{I}_{k}$ and then linearly interpolate the RGB colors of neighborhood pixels.
}

We also consider the measurement noise of $\boldsymbol{\gamma}_s$ and $\mathbf{c}_s$: 
\begin{align}
\boldsymbol{\gamma}_s  =  \boldsymbol{\gamma}^{gt}_s + \mathbf{n}_{{\boldsymbol{\gamma}_s}}, ~&  \mathbf{n}_{{\boldsymbol{\gamma}_s}} \sim \mathcal{N}( \mathbf{0}, \boldsymbol{\Sigma}_{\mathbf{n}_{{\boldsymbol{\gamma}_s}}}   ) \label{eq_measurement_noise_gamma_s} \\
\begin{split}
	\mathbf{c}_s = \mathbf{c}_s^{gt} + \mathbf{n}_{\mathbf{c}_s} +  \boldsymbol{\eta}_{\mathbf{c}_s} \label{eq_measurement_noise_C_s} , ~& \mathbf{n}_{\mathbf{c}_s} \sim {\mathcal{N}}(\mathbf{0}, \boldsymbol{\Sigma}_{\mathbf{n}_{\mathbf{c}_s}} ) \\  & 
	\hspace{-1.2cm}\boldsymbol{\eta}_{\mathbf{c}_s} \sim  {\mathcal{N}}(\mathbf{0},  \boldsymbol{\sigma}^2_{s} \cdot \Delta t_{\mathbf{c}_s} )
\end{split}
\end{align}
where $\boldsymbol{\gamma}^{gt}_s$ is the ground true value of $\boldsymbol{\gamma}_s$ and $\mathbf{c}_s^{gt}$  is the true value of $\mathbf{c}_s$, $ \Delta t_{\mathbf{c}_s}$  is the time internal between current frame and last rendering time of $\mathbf{P}_s$. Notice that in (\ref{eq_measurement_noise_C_s}), the measurement noise of $\mathbf{c}_s$ consists of two parts: the estimated error  $\mathbf{n}_{\mathbf{c}_s}$ from the last time of rendering (see Section. \ref{sect_render_texture_of_maps}), and the impulsed random walk process noise $\boldsymbol{\eta}_{\mathbf{c}_s}$, which accounts for the change of environment illusion.

%Even though $\boldsymbol{\eta}_{\mathbf{c}_s}$ in (\ref{eq_measurement_noise_C_s}) might not the best precise way that modeling the errors, we hold the view that it is the cheapest solution we can adopt in this work.

Combining (\ref{eq_frame_to_map_photo_err}), (\ref{eq_measurement_noise_gamma_s}) and (\ref{eq_measurement_noise_C_s}), we obtain the first order Taylor expansion of the true zero residual $\mathbf{o}( \mathbf{x}_k, {^G}\mathbf{p}^{gt}_s, \mathbf{c}^{gt}_s )$:
\begin{align}
\begin{split}
	\mathbf{0} &= \mathbf{o}( \mathbf{x}_k, {^G}\mathbf{p}^{gt}_s, \mathbf{c}^{gt}_s ) \approx \mathbf{o}( \check{\mathbf{x}}_k, {^G}\mathbf{p}_s, \mathbf{c}_s ) + \mathbf{H}_{s}^{o} \delta \check{\mathbf{x}}_k + \boldsymbol{\beta}_{s} \label{eq_visual_pe_residual}
\end{split}
\end{align} 
where $\boldsymbol{\beta}_{s} \sim \mathcal{N}(\mathbf{0}, \boldsymbol{\Sigma}_{\beta_{s}})$ and:
\begin{align}
	\hspace{-0.2cm}\mathbf{H}^{o}_{s}  &=\left.\dfrac{ \partial  \mathbf{o}( \check{\mathbf{x}}_{k} \boxplus \delta{\check{\mathbf{x}}}_{k},   {^G}\mathbf{p}_s, \mathbf{c}_s ) }{\partial  \delta{\check{\mathbf{x}}}_{k}} \right|_{\delta{\check{\mathbf{x}}}_{k} = \mathbf{0}}  \\
	{\boldsymbol{\Sigma}_{\boldsymbol{\beta}_{s}}} &= \boldsymbol{\Sigma}_{\mathbf{n}_{\mathbf{c}_s}} + \boldsymbol{\sigma}^2_{s} \cdot \Delta t_{\mathbf{c}_s} +  \boldsymbol{\Sigma}_{\mathbf{n}_{{\boldsymbol{\gamma}_s}}}  + {\mathbf{F}^o_{{{\mathbf{p}}}_{s}}} \boldsymbol{\Sigma}_{\mathbf{p}_{s}} {\mathbf{F}^o_{{{\mathbf{p}}}_{s}}}^T \label{eq_def_H_s} \\
	{\mathbf{F}^o_{{{\mathbf{p}}}_{s}}} &=   
	\dfrac{ \partial  \mathbf{o}(\check{\mathbf{x}}_{k}, {^G}\mathbf{p}_s, \mathbf{c}_s)  }{\partial  {^{G}{\mathbf{p}}}_{s} } 
\end{align}

%For the detail computation of $\mathbf{H}^{o}_{s}$ and $\mathbf{F}^o_{{{\mathbf{p}}}_{s}}$, please refer our detail derivation in supplemental material.\todo{XXX}.

\subsubsection{Frame-to-map VIO ESIKF update}\label{sect_frame_to_map_ESIKF_update}
Equation (\ref{eq_visual_pe_residual}) constitutes another observation distribution for $\delta \check{\mathbf{x}}_k$, which is combined with the prior distribution from the IMU propagation to obtain the maximum a posteriori (MAP) estimation of $\delta \check{\mathbf{x}}_{k}$:
	\begin{align}
	\begin{split}
		%       \begin{split}
		\hspace{-0.5cm}\mathop{\min}_{\delta \check{\mathbf{x}}_{k}} &\left(  \left\| \check{\mathbf{x}}_{k} \boxminus \hat{\mathbf{x}}_{k}  + \boldsymbol{\mathcal{H}} \delta{\check{\mathbf{x}}}_{k}   \right\|_{\boldsymbol{\Sigma}_{\delta \hat{\mathbf{x}}_{k}} }^2 \right.  \\
		+&\left. \sum\nolimits_{s=1}^{m}\left\|  {  \mathbf{o}( \check{\mathbf{x}}_k, {^G}\mathbf{p}_s, \mathbf{c}_s ) + \mathbf{H}^{o}_{s} \delta \check{\mathbf{x}}_{k} } \right\|^2_{{\boldsymbol{\Sigma}_{\boldsymbol{\beta}_{s}}}}  \right) 
		%       \end{split}
		\label{eq_optimial_frame_to_map}
	\end{split}
\end{align}

Denote $\mathbf{H}, \mathbf{R}, \check{\mathbf{z}}_k$ and $\mathbf{P}$ similar to (\ref{eq_H_mat_frame_to_frame}) $\sim$ (\ref{eq_P_mat_frame_to_frame}) :  
%\begin{small}
\begin{align}
	\mathbf{H} &= 
	\begin{bmatrix}
		{\mathbf{H}^{o}_{1}}^T ,\dots, {\mathbf{H}^{o}_{m}}^T 
	\end{bmatrix}^T\\
	\mathbf{R} &= 
	\text{diag}(
	\begin{matrix}
		\boldsymbol{\Sigma}_{\boldsymbol{\beta}_{1}} , \dots ,\boldsymbol{\Sigma}_{\boldsymbol{\beta}_{m}}
	\end{matrix}
	)\\
	\check{\mathbf{z}}_{k} &= 
	\left[
	\mathbf{o}( \check{\mathbf{x}}_k, {^G}\mathbf{p}_1, \mathbf{c}_1 ) \dots,  
	\mathbf{o}( \check{\mathbf{x}}_k, {^G}\mathbf{p}_m, \mathbf{c}_m ) \right] ^T  \\
	\mathbf{P} &= 
	\left(\boldsymbol{\mathcal{H}}\right)^{-1} \boldsymbol{\Sigma}_{\delta \hat{\mathbf{x}}_{k}} {\left(\boldsymbol{\mathcal{H}}\right)}^{-T}
\end{align}
%\end{small}
then we perform the state update similar to (\ref{eq_K_mat_frame_to_frame}) and (\ref{eq_x_k_update_frame_to_frame}). 
This frame-to-map VIO ESIKF update (Section \ref{sect_frame_to_map_PE} to Section \ref{sect_frame_to_map_ESIKF_update}) is iterated until convergence. The converged state estimate is then used to: (1) render the texture of maps (Section \ref{sect_render_texture_of_maps});
(2) update the current tracking point set  $\boldsymbol{\mathcal{P}}$ for the next frame to use (Section \ref{sect_update_tracking_pts}); and (3) serve as the starting point of the IMU propagation in the next frame of LIO or VIO update.
\begin{align}
	\hat{\mathbf{x}}_{k} = \check{\mathbf{x}}_{k}, \hspace{0.2cm}
	\boldsymbol{\Sigma}_{\delta \hat{\mathbf{x}}_{k}} =  \left( \mathbf{I}- \mathbf{KH} \right)\boldsymbol{\Sigma}_{\delta \check{\mathbf{x}}_{k}}
\end{align}

\subsection{Render the texture of global map}\label{sect_render_texture_of_maps} 
\begin{figure}[t]
	\centering
	\includegraphics[width=1.0\linewidth]{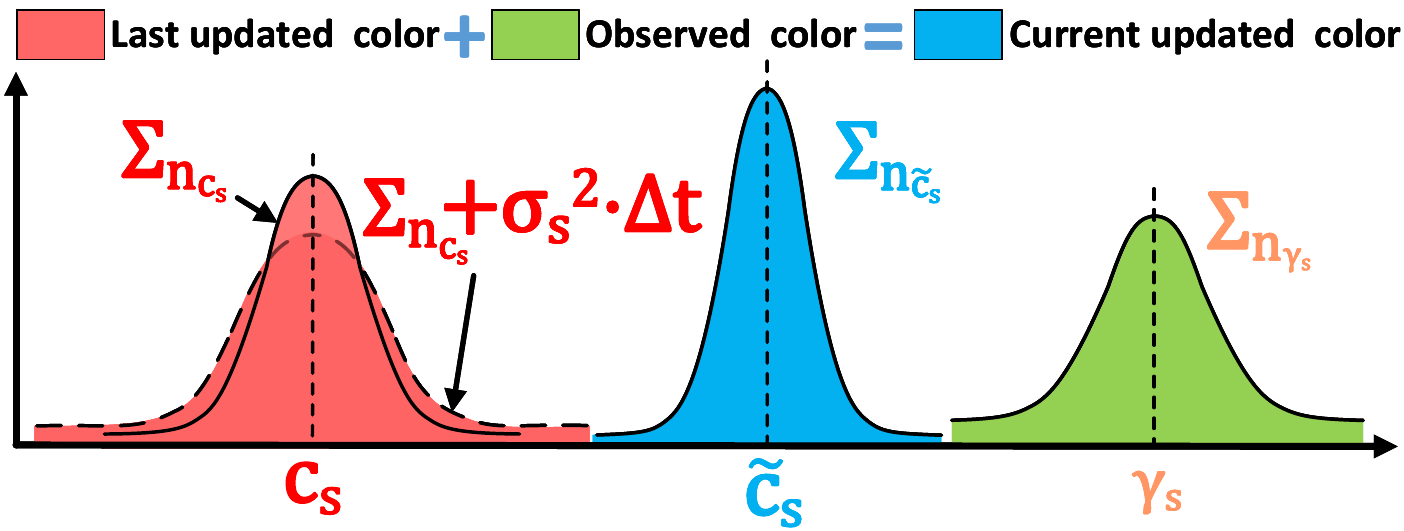}
	\caption{We render the color of point $\mathbf{c}_s$ via Bayesian update.}
	\label{fig_color_update}
\end{figure}
After the frame-to-map VIO update, we have the precise pose of the current image, and then we perform the rendering function to update the color of map points.

First of all, we retrive all the points in all \textit{activated} voxels (see Section. \ref{sect_LIO_subsystem}). Assume the total number points is $n$ points and the point set $\boldsymbol{\zeta} = \{ \mathbf{P}_1, ..., \mathbf{P}_n \}$. Taking the color update process of the $s$-th point $\mathbf{P}_s = \left[ {^G}\mathbf{p}_s^T, \mathbf{c}_s^T \right]^T \in \boldsymbol{\zeta}$ as an example. If $\mathbf{P}_s$ falls in the current image $\mathbf{I}_k$, we obtain its observation color $\boldsymbol{\gamma}_s$ and covariance $\boldsymbol{\Sigma}_{\mathbf{n}_{{\boldsymbol{\gamma}_s}}}$ by linearly interpolating the RGB values of the neighborhood pixels on $\mathbf{I}_k$. The newly observed point color is fused with the existing color $\mathbf{c}_s$ recorded in the map via Bayesian update (see Fig. \ref{fig_color_update}), leading to the updated color and the associated covariance of $\mathbf{c}_s$ as follows:
%\begin{small}
\begin{align}
	\hspace{-0.5cm}\boldsymbol{\Sigma}_{\mathbf{n}_{\tilde{\mathbf{c}}_s}} = \left( \left(\boldsymbol{\Sigma}_{\mathbf{n}_{\mathbf{c}_s}} + \boldsymbol{\sigma}^2_{s} \cdot \Delta t_{\mathbf{c}_s} \right)^{-1} + \boldsymbol{\Sigma}_{\mathbf{n}_{\boldsymbol{\gamma}_s}}^{-1} \right)^{-1} \\
	\hspace{-0.5cm}\tilde{\mathbf{c}}_{s} = \left( \left(\boldsymbol{\Sigma}_{\mathbf{n}_{\mathbf{c}_s}} + \boldsymbol{\sigma}^2_{s} \cdot \Delta t_{\mathbf{c}_s} \right)^{-1} \mathbf{c}_s + \boldsymbol{\Sigma}_{\mathbf{n}_{\boldsymbol{\gamma}_s}}^{-1}  \boldsymbol{\gamma}_s \right)^{-1} \boldsymbol{\Sigma}_{\mathbf{n}_{\tilde{\mathbf{c}}_s}} \\
	 \hspace{1.5cm}{\mathbf{c}}_{s} = \tilde{\mathbf{c}}_{s}, ~~ {\boldsymbol{\Sigma}}_{\mathbf{n}_{\mathbf{c}_s}} = \boldsymbol{\Sigma}_{\mathbf{n}_{\tilde{\mathbf{c}}_s}} \hspace{1.5cm}
\end{align} 
%\end{small}

\subsection{Update of the tracking points of VIO subsystem}\label{sect_update_tracking_pts}
After the rendering of texture, we perform the update of tracked point set $\boldsymbol{\mathcal{P}}$. Firstly, we remove the points from $\boldsymbol{\mathcal{P}}$ if their  projection error in (\ref{eq_frame_to_map_projection_err}) or photometric error in (\ref{eq_frame_to_map_photo_err}) are too large, and also remove the points which does not fall into $\mathbf{I}_k$. Secondly, we project each point in $\boldsymbol{\zeta}$ to the current image $\mathbf{I}_k$, and add it to $\boldsymbol{\mathcal{P}}$ if there have no other tracked points located nearby (e.g. in the radius of $50$ pixels).

\section{Experiment and Results}
\subsection{Equipment setup}
\begin{figure}[t]
	\centering
	\includegraphics[width=1.0\linewidth]{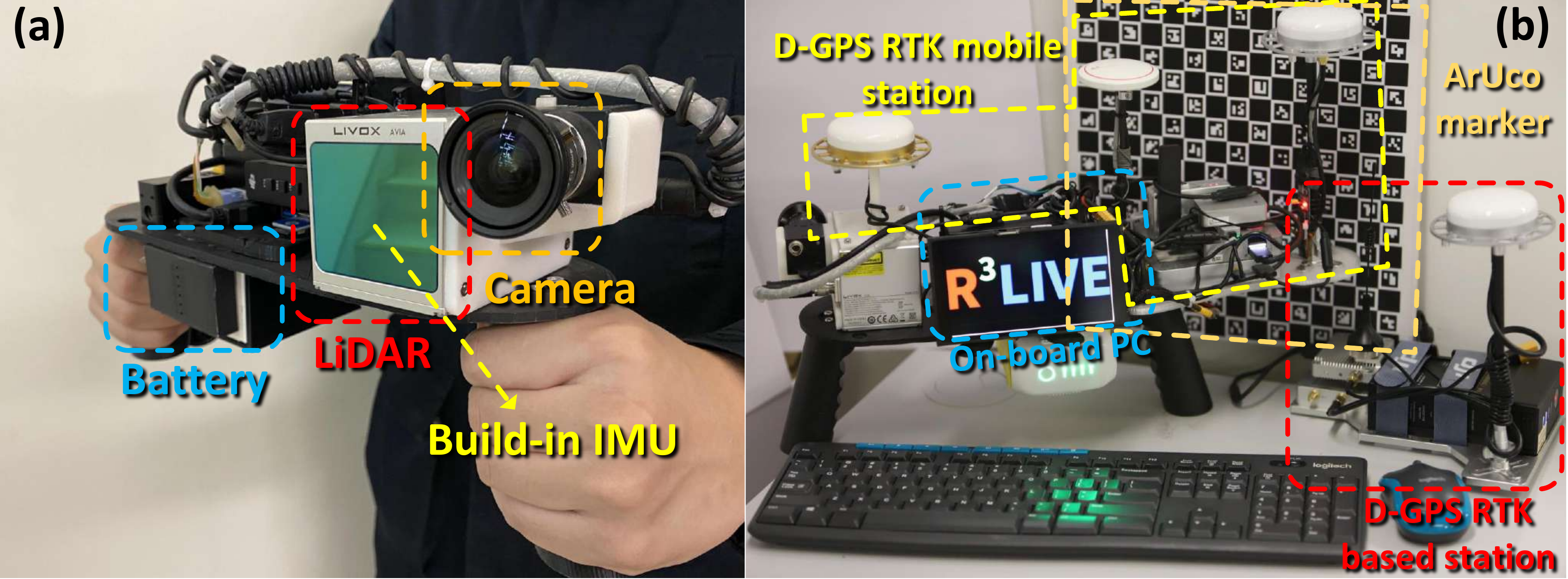}
	\caption{Our handheld device for data collection,  (a) shows our minimum system, with a total weight of $2.09$ Kg; (b) an additional D-GPS RTK system and an ArUco marker board \cite{garrido2014automatic} are used to evaluate the system accuracy.}
	\vspace{-0.8cm}
	\label{fig_handheld_device}
\end{figure}

Our handheld device for data collection is shown in Fig. \ref{fig_handheld_device} (a), which includes a power supply unit, the onboard DJI \textit{manifold-2c}\footnote{\url{https://www.dji.com/manifold-2}} computation platform (equipped with  an \textit{Intel i7-8550u} CPU and 8 GB RAM), a \textit{FLIR Blackfly BFS-u3-13y3c} global shutter camera,  and a \textit{LiVOX AVIA}\footnote{\url{https://www.livoxtech.com/avia}} LiDAR. The FoV of the camera is $82.9^\circ \times 66.5 ^\circ$ while the FoV of the LiDAR is $70.4^\circ \times 77.2 ^\circ$. To quantitatively evaluate the precision of our algorithm (Section \ref{sect_experiment_3}), we install a Differential-GPS (D-GPS) real-time kinematic (RTK) system\footnote{\url{https://www.dji.com/d-rtk}} on our device, shown in Fig. \ref{fig_handheld_device} (b). In GPS denied environment (Section \ref{sect_experiment_1} and Section \ref{sect_experiment_2}), we use the ArUco marker \cite{garrido2014automatic} for calculating the drift of our odometry. Notice that all of the mechanical modules of this device are designed as FDM printable\footnote{\url{https://en.wikipedia.org/wiki/Fused\_filament\_fabrication}}, and to faciliate our reader to reproduce our work, we open source all of our schematics on our github\footnote{\url{https://github.com/ziv-lin/rxlive_handheld}}.

\subsection{Experiment-1: Robustness evaluation in simultaneously LiDAR degenerated and visual texture-less environments}\label{sect_experiment_1}

\begin{figure}[h]
	\centering
	\includegraphics[width=1.0\linewidth]{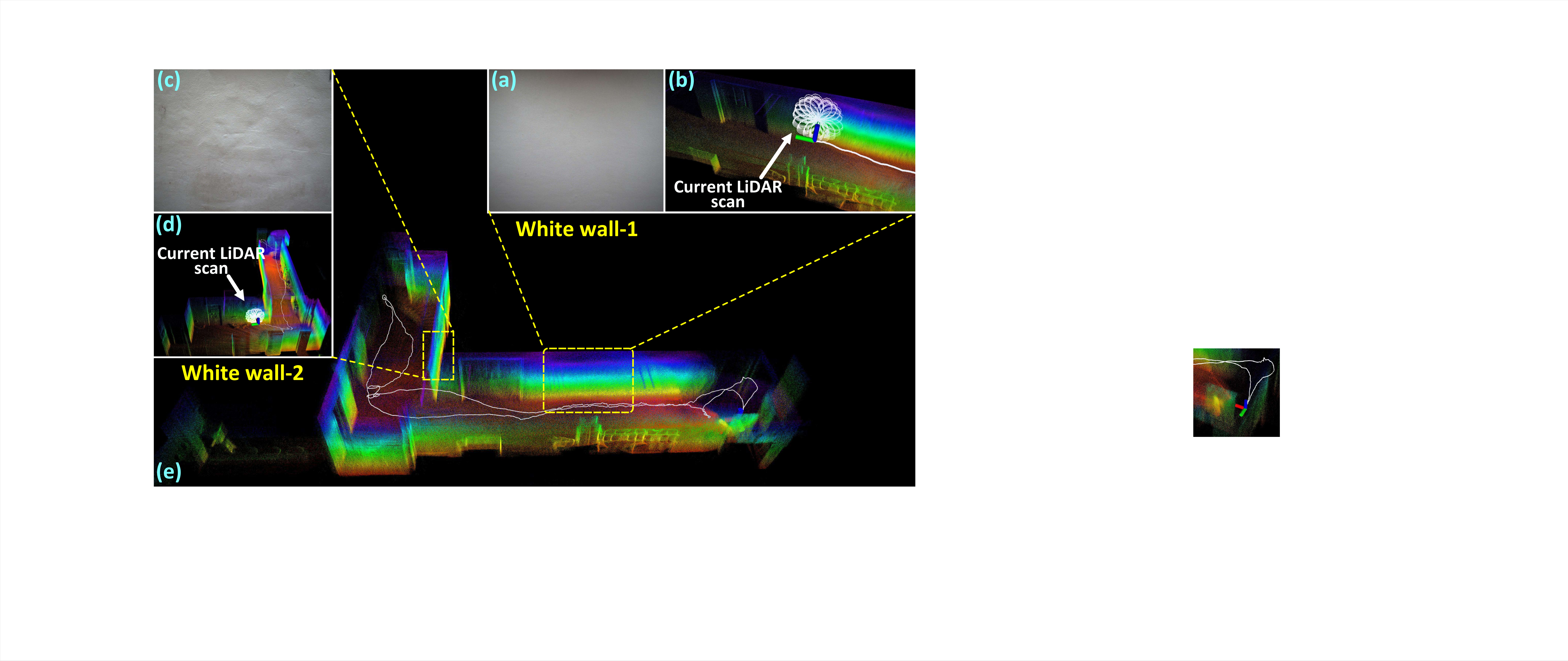}
	\caption{We test our algorithm in simultaneously LiDAR degenerated and visual texture-less scenarios.}
	\label{fig_exp_1}
	\includegraphics[width=1.0\linewidth]{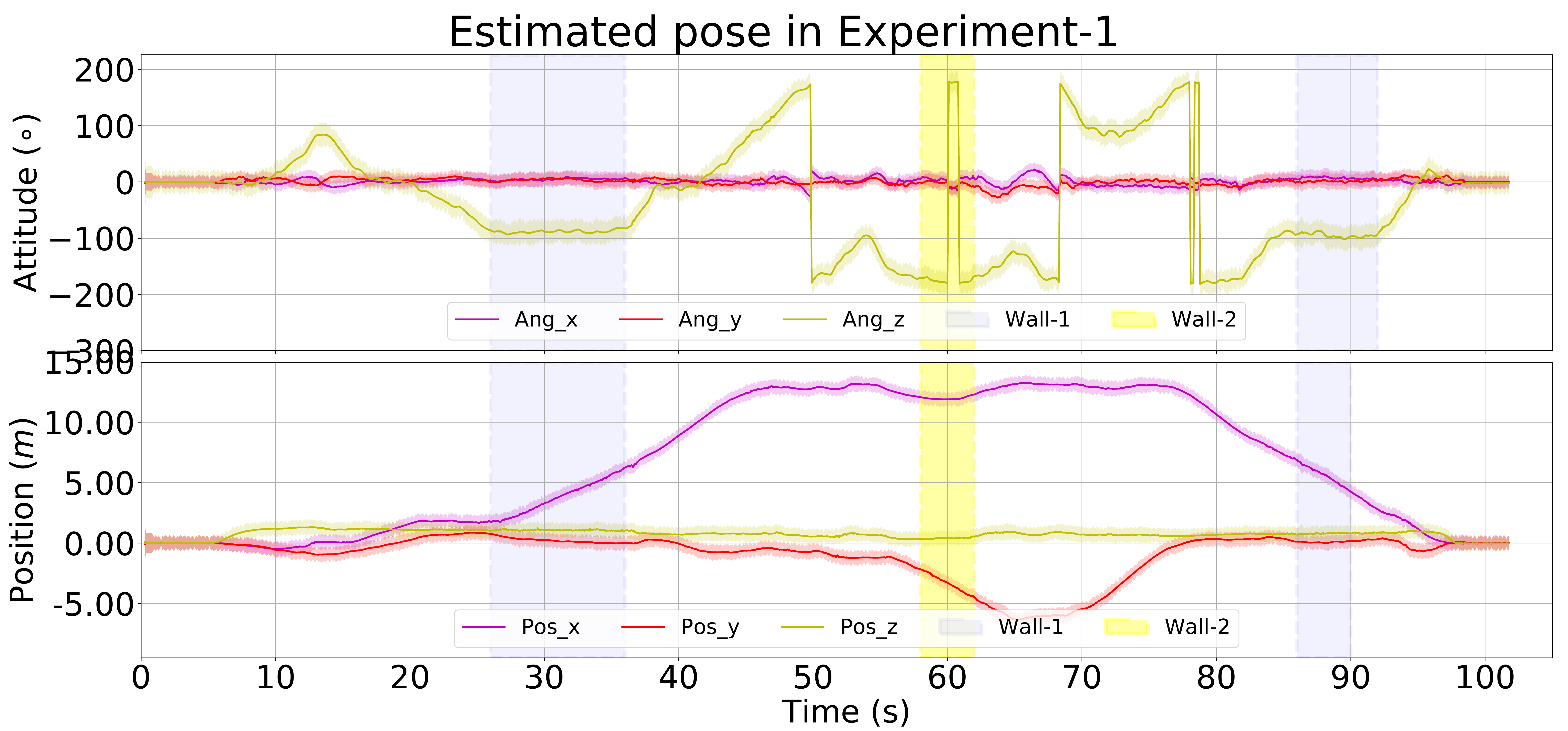}
	\caption{The estimated pose and its 3-$\sigma$ bound with 5 times amplification for better visualization (the light-colored area around the trajectory) of Experiment-1. The shaded areas in blue and yellow are the phases of the sensors facing against the white ``wall-1" and ``wall-2", respectively. In the end, the algorithm drifts  $1.62 ^\circ$ in rotation and $4.57 $ cm in translation.}
	\label{fig_exp_1_pose}
	\vspace{-0.5cm}
\end{figure}

In this experiment, we challenge one of the most difficult scenarios in SLAM, we perform the robustness evaluation in simultaneously LiDAR degenerated and visual texture-less environments. As shown in Fig. \ref{fig_exp_1}, our sensors pass through a narrow ``T"-shape passage while occasionally facing against the side walls. When facing the wall which imposes only a single plane constraint, the LiDAR is well-known degenerated for full pose estimation. Meanwhile, the visual texture on the white walls are very limited (Fig. \ref{fig_exp_1}(a) and Fig. \ref{fig_exp_1}(c)), especially for the wall-1, which has the only changes of illumination. Such scenarios are challenging for both LiDAR-based and visual-based SLAM methods. 

Taking advantage of using the raw image pixel color information, our proposed algorithm can ``survive" in such 
extremely difficult scenarios. Fig. \ref{fig_exp_1_pose} shows our estimated pose, with the phases of passing through ``wall-1" and ``wall-2" shaded with blue and yellow, respectively. The estimated covariance is also shown in Fig. \ref{fig_exp_1_pose}, which are bounded over the whole estimated trajectory, indicating that our estimation quality is stable over the whole process. The sensor is moved to the starting point, where an ArUco marker board is used to obtain the ground-truth relative pose between the starting and end poses. Compared with the ground-truth end pose, our algorithm drifts  $1.62 ^\circ$ in rotation and $4.57 $ cm in translation. We recommend the readers to the accompanying video on
YouTube\footnote{\url{https://youtu.be/j5fT8NE5fdg?t=6}} for better visualization of the experiment.
 
\begin{figure*}[t]
	\centering
	\includegraphics[width=1.0\linewidth]{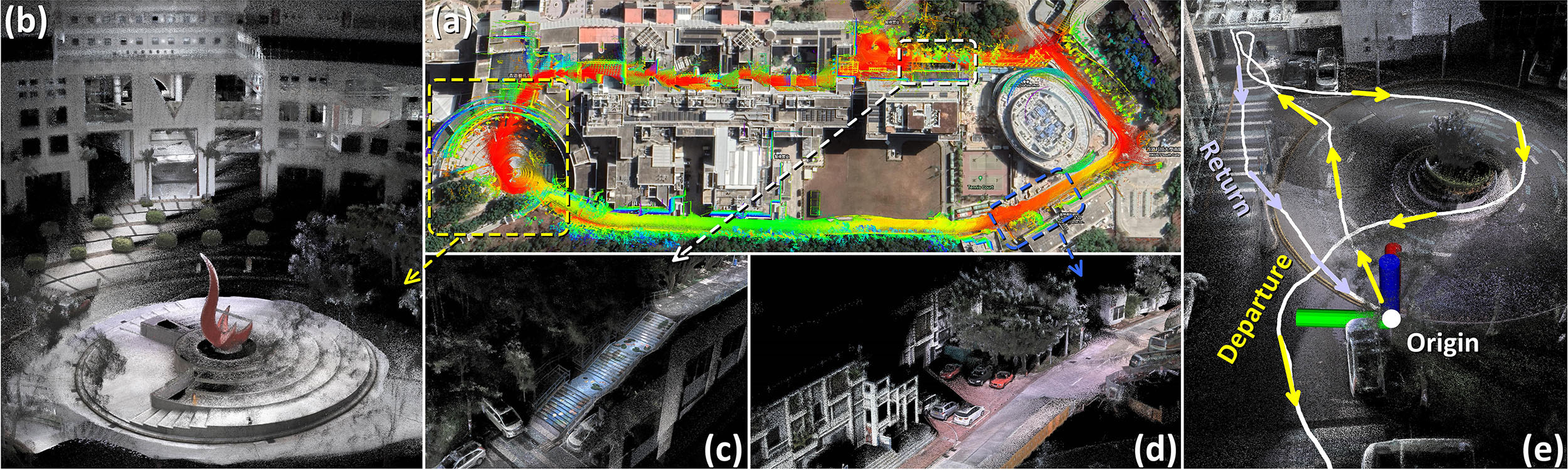}
	\caption{We show the dense 3D RGB-colored maps reconstructed in real-time in Experiment-2. In (a), we merge our maps with the Google-Earth satellite image and find them aligned well. Details of the map are selectively shown in (b), (c), and (d), respectively.  In (e), we show that our algorithm can close the loop itself without any additional processing (e.g. loop-detection and closure), with the return trajectory can return back to the origin. We refer readers to the video on our Youtube\protect\footnotemark for more details.}
	\label{fig_exp2_maps}
	\vspace{-0.7cm}
\end{figure*}
\footnotetext{\url{https://youtu.be/j5fT8NE5fdg?t=104}}

\subsection{Experiment-2: High precision mapping large-scale indoor \& outdoor urban environment}\label{sect_experiment_2}

\begin{figure}[h]
	\centering
	\includegraphics[width=0.8\linewidth]{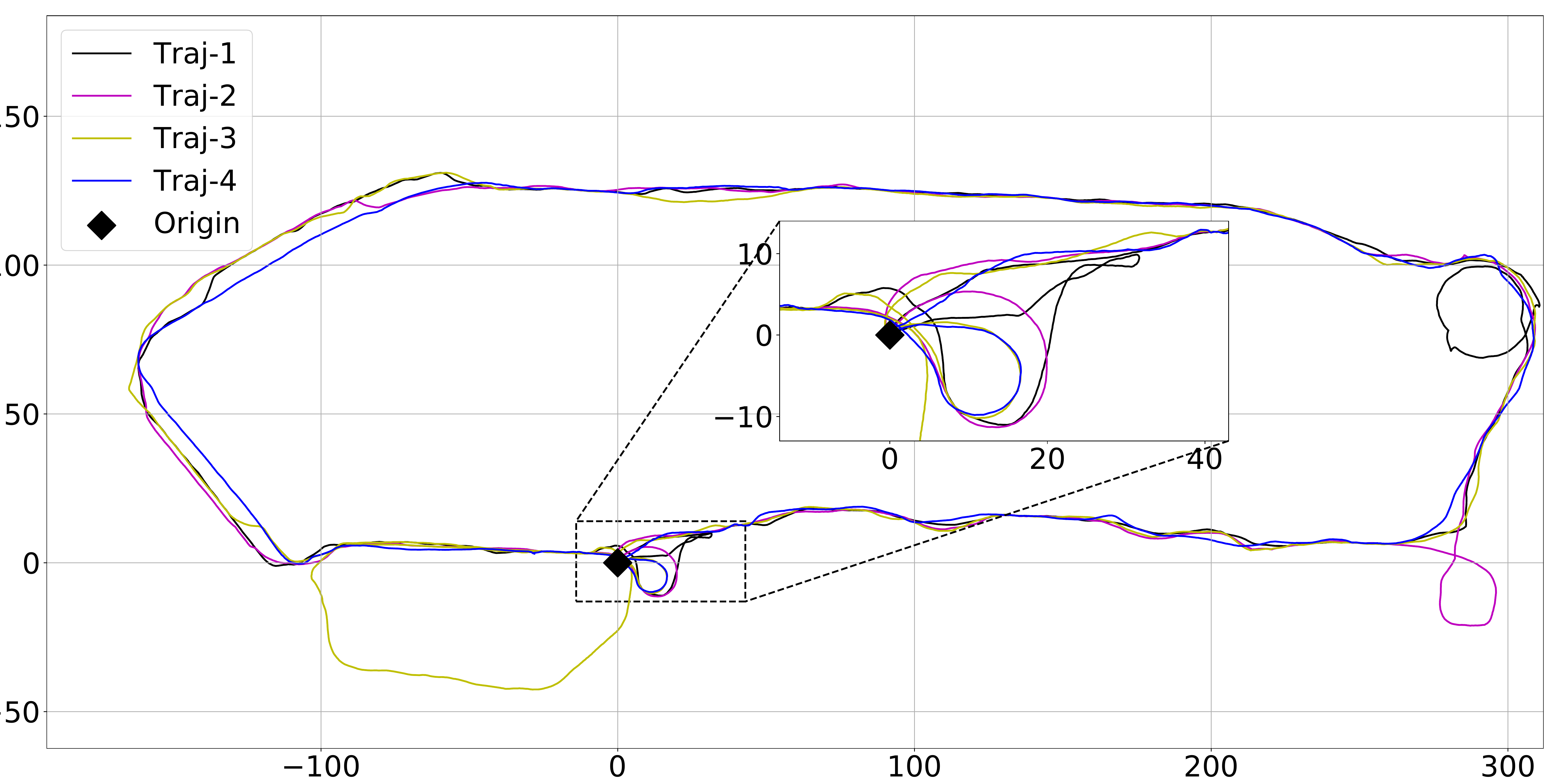}
	\caption{The birdview of the 4 trajectories in Experiments-2, the total length of the four trajectories are $1317$ m , $1524$ m, $1372$ m and $1191$ m, respectively.}
	\label{fig_exp2_birdview}
	\includegraphics[width=0.8\linewidth]{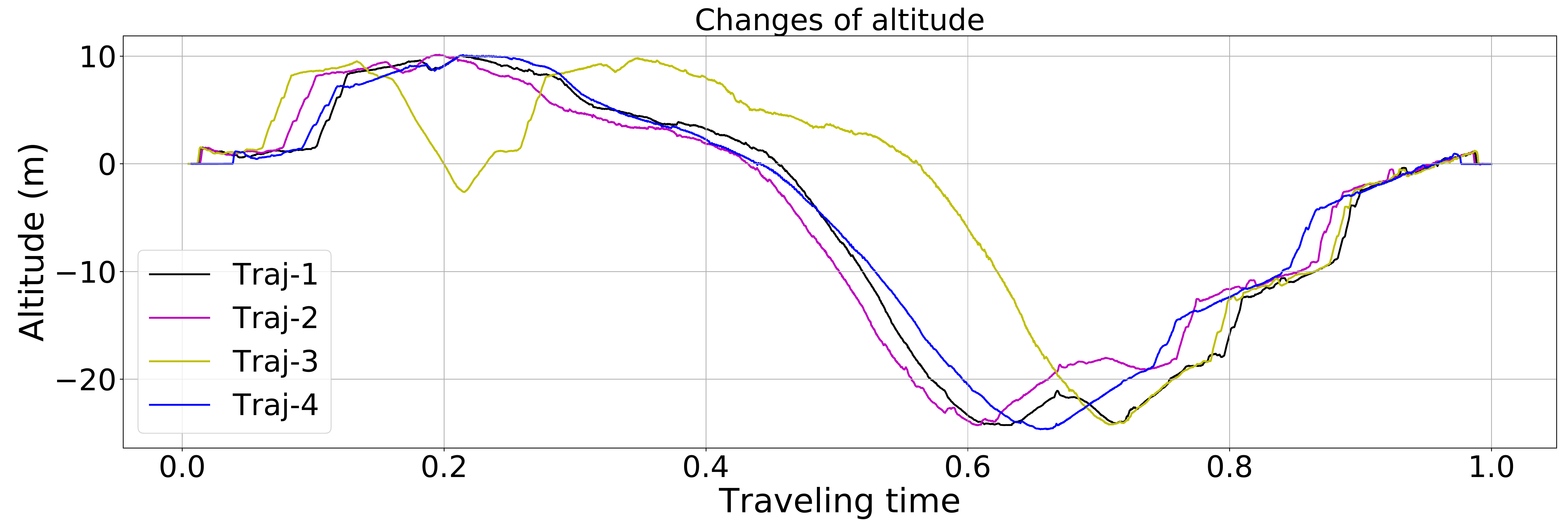}
	\caption{The changes of altitude among the 4 trajectories in Experiment-2, where the time are normalized to 1.}
	\label{fig_exp2_altitude}
	%	\vspace{-0.3cm}
\end{figure}
In this experiment, we show the capacity of our algorithm for reconstructing a precise,  dense, 3D, RGB-colored map of a large-scale campus environment in real-time. We collect the data within the campus of the Hong Kong University of Science and Technology (HKUST) 4 times with different traveling trajectories (i.e. Traj 1$\sim $4), with their total length of $1317$, $1524$, $1372$ and $1191$ meters, respectively. The bird-view (i.e. project on X$-$Y plane) of these trajectories are shown Fig. \ref{fig_exp2_birdview} and the changes of their altitude are plotted in Fig. \ref{fig_exp2_altitude}. Without any additional processing (e.g. loop closure), all of these four trajectories can close the loop itself (see Fig.\ref{fig_exp2_maps} (e)). Using the ArUco marker board placed at the starting point, the odometry drifts are presented in Table. \ref{Table_I_drift}, which indicates that our proposed method is of high accuracy with small drifts over the long trajectories and in complicated environments. Finally, we present the the reconstructed map in ``Traj-1'' in Fig. \ref{fig_exp2_maps}. More visualization results are available on the project page\footnote{\url{https://github.com/ziv-lin/r3live_preview}}.  

\begin{table}[h]
	\centering
	%\captionsetup{justification=centering}
	\caption{Odometry drift in 4 trajectories of Experiment-2.}
	\begin{tabular}{|c|c|c|c|c|}
		\hline
		& Traj-1 & Traj-2 & Traj-3 & Traj-4 \\ \hline
		Length of trajectory (m)            & 1317   & 1524   & 1372   & 1191   \\ \hline
		Drift in translation (m) & 0.093  & 0.154  & 0.164  & 0.102  \\ \hline
		Drift in rotation (deg)  & 2.140   & 0.285   & 2.342   & 3.925   \\ \hline
	\end{tabular}
	\label{Table_I_drift}
	\vspace{-0.3cm}
\end{table}

\subsection{Experiment-3: Quantitative evaluation of precision using D-GPS RTK}\label{sect_experiment_3}
\begin{figure}[h]
	\centering
	{
		\includegraphics[width=0.49\linewidth]{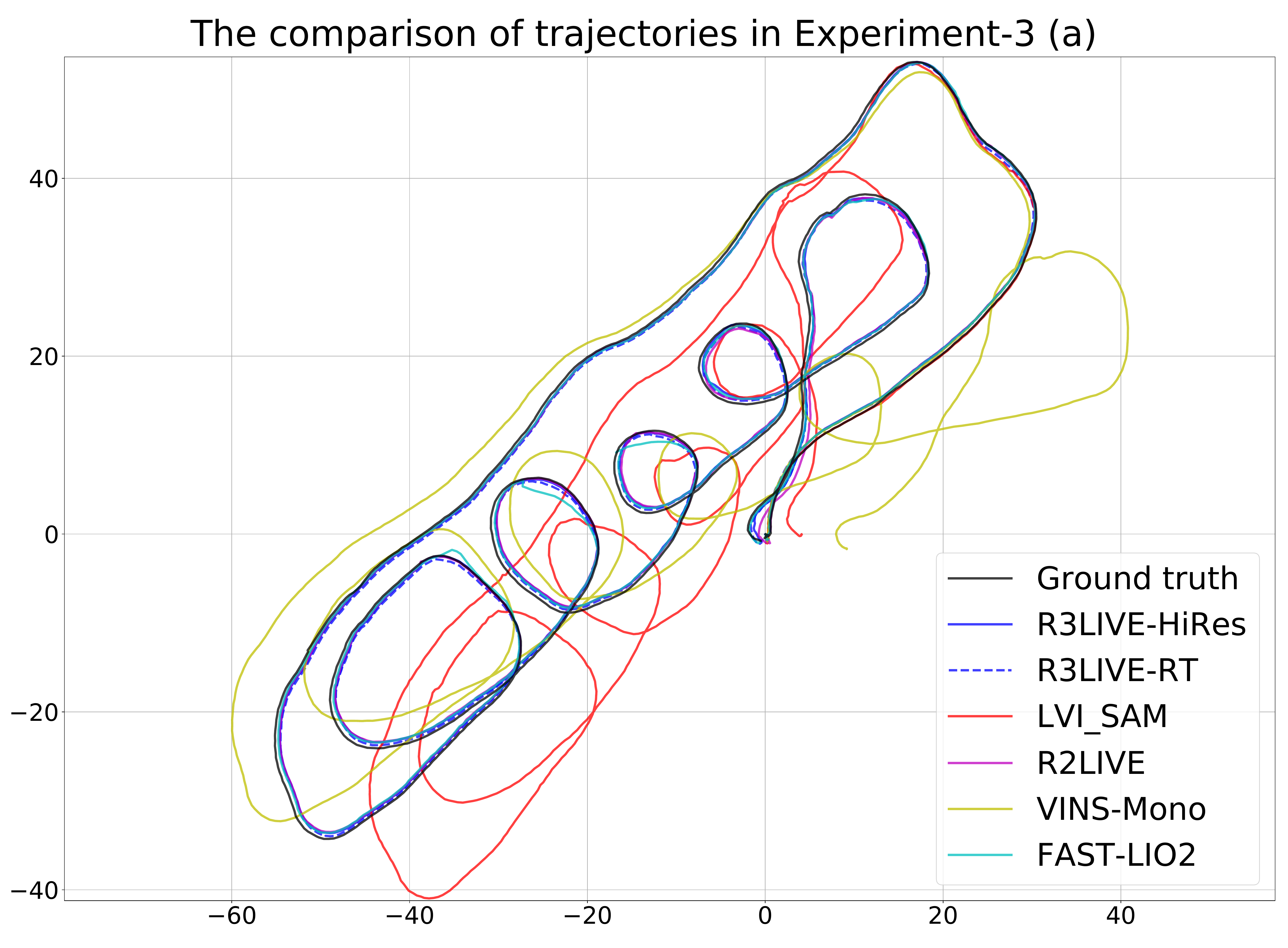}
		\includegraphics[width=0.49\linewidth]{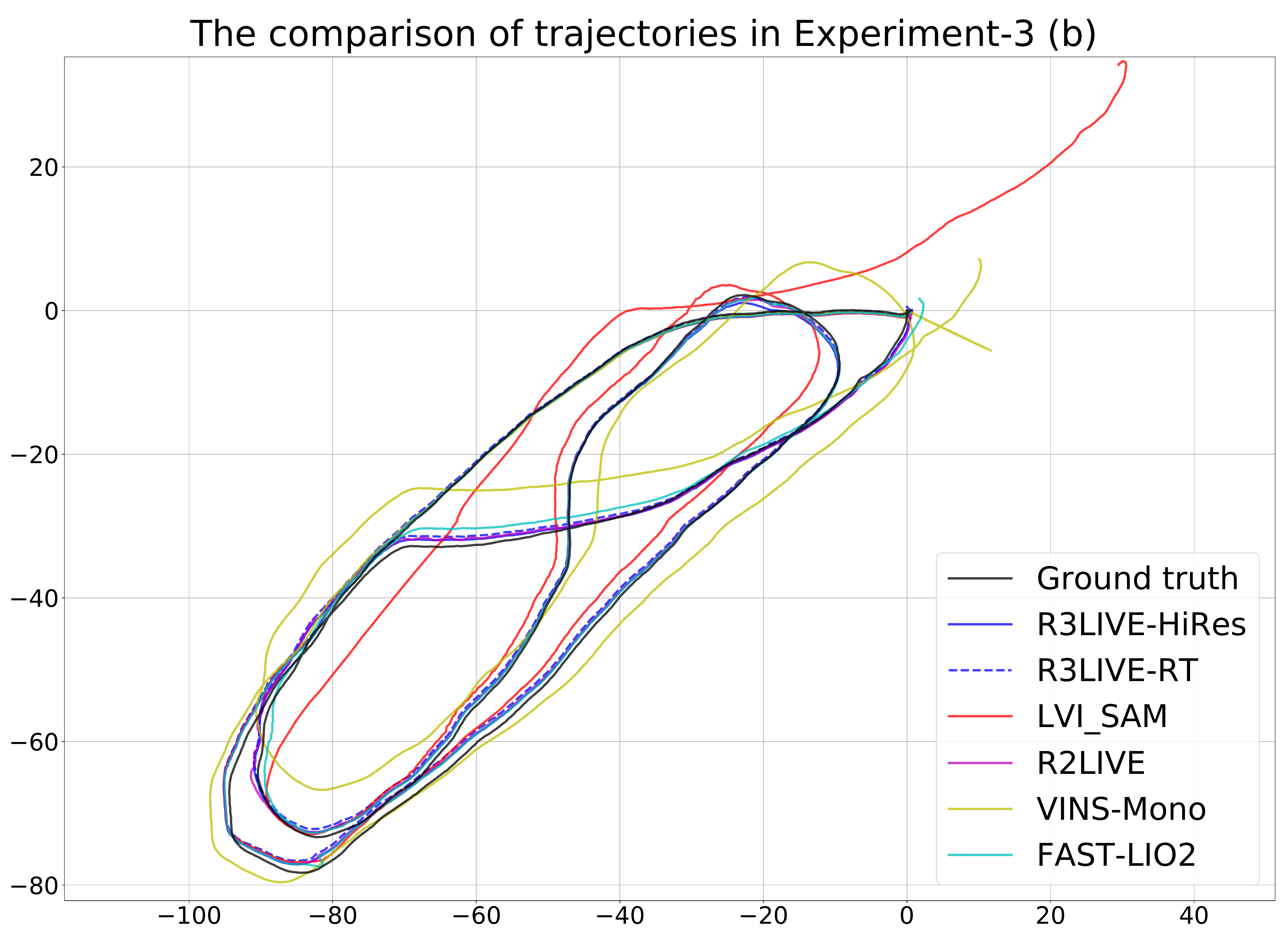}
		\caption{Comparison of the estimated trajectories in Experiments-3.}
		\label{fig_eval_rtk}
		\vspace{-0.3cm}
	}
%	\caption{The changes of altitude among 4 trajectories in our Experiment-2, where the time are normalized to 1. \todo{Change the legend ``Traj-X" and the label size}}	
%	\label{fig_gps_04}
%	\caption{Experiment 2.}
	%	\vspace{-0.3cm}
\end{figure}

In this experiment, we collect two sequences of data in a seaport (\textit{Belcher Bay Promenade}\footnote{\url{https://goo.gl/maps/eikg6Kvg9k4gM3Sm9}}) with a real-time Differential-GPS (D-GPS) Kinematic (RTK) system providing the ground-truth trajectory. In these two sequences, the sensors often faces many pedestrians and occasionally open seas, where the LiDAR has very few measurements.

 We compare the trajectories estimated by R$^3$LIVE with two different configurations (``R3LIVE-HiRES" and ``R3LIVE-RT", see Table. \ref{table_experiments_precision}), ``LVI-SAM" (with modified its LiDAR front-end for Livox Avia LiDAR),  ``R$^2$LIVE"\cite{r2live}, ``VINS-Mono" (IMU+Camera)\cite{qin2018vins}, ``Fast-LIO2" (IMU+LiDAR)\cite{fastlio2} with the ground-truth in Fig. \ref{fig_eval_rtk}, where we can see that our estimated trajectories agree the best with the ground-truth in both sequences. To have more quantitative comparisons, we compute the relative rotation error (RPE) and relative translation error (RTE) \cite{zhang2018tutorial} over all possible sub-sequences of length (50,100,150,...,300) meters are shown in Table. \ref{table_experiments_precision}.

\begin{table}[]
	\scriptsize
	\caption{The relative pose error in Experiment-3. In configuration ``R3LIVE-HiRES", we set the input image resolution as $1280 \times 1024$ and the minimum distance between two points in maps as $0.01$ meters. In configuration ``R3LIVE-RT", we set the input image resolution as $320 \times 256$ and the minimum distance between two points in maps as $0.10$ meters }
	\setlength\tabcolsep{1.0pt}
	\label{table_experiments_precision}	
	\begin{tabular}{|c|c|c|c|c|c|c|}
		\hline
	\multicolumn{7}{|c|}{\scriptsize Relative pose error in Experiments-3 (a)}         \\ \hline
	Sub-sequence & 50 m         & 100 m       & 150 m       & 200 m       & 250 m       & 300 m       \\
	RRE/RTE      & deg / \%     & deg / \%    & deg / \%    & deg / \%    & deg / \%    & deg / \%    \\ \hline
	R3LIVE-HiRes  & \textbf{0.99} / 2.25  & \textbf{0.53} / \textbf{1.02} & \textbf{0.46} / \textbf{0.60} & \textbf{0.29} / \textbf{0.31} & \textbf{0.24} / 0.31 & \textbf{0.21} / \textbf{0.17} \\ \hline
	R3LIVE-RT    & 1.48 / \textbf{2.21}  & 0.54 / 1.06 & 0.49 / 0.60 & 0.31 / 0.41 & 0.25/ \textbf{0.31}  & 0.22 / 0.23 \\ \hline
	LVI\_SAM     & 2.11 / 13.73 & 1.04 / 6.69 & 0.83 / 5.07 & 0.60 / 3.86 & 0.47 / 2.98 & 0.43 / 2.40 \\ \hline
	R2LIVE       & 1.21 / 2.47  & 0.61 / 1.02 & 0.59 / 0.60 & 0.34 / 0.34 & 0.37 / 0.33 & 0.34 / 0.21 \\ \hline
	FAST-LIO2    & 1.36 / 2.35  & 0.66 / 0.82 & 0.47 / 0.60 & 0.37 / 0.36 & 0.37 / 0.31 & 0.21 / 0.20 \\ \hline
	VINS-Mono    & 3.03 / 10.41 & 1.70 / 7.63 & 1.15 / 6.36 & 0.89 / 4.34 & 0.73 / 3.08 & 0.59 / 2.31 \\ \hline
%	R3LIVE-LIO   & 2.04 / 2.49  & 1.01 / 0.86 & 0.67 / 0.59 & 0.61 / 0.38 & 0.50 / 0.33 & 0.44 / 0.22 \\ \hline
	\multicolumn{7}{|c|}{\scriptsize Relative pose error in Experiments-3 (b)}         \\ \hline
	Sub-sequence & 50 m         & 100 m       & 150 m       & 200 m       & 250 m       & 300 m       \\
	RRE/RTE      & deg / \%     & deg / \%    & deg / \%    & deg / \%    & deg / \%    & deg / \%    \\ \hline
	R3LIVE-HiRes  & 1.06 / \textbf{1.98}  & 0.58 / \textbf{1.34} & \textbf{0.43} / \textbf{0.83} & \textbf{0.32} / \textbf{0.59} & 0.27 / \textbf{0.27} & 0.25 / 0.16 \\ \hline
	R3LIVE-RT    & \textbf{0.98} / 2.08  & \textbf{0.55} / 1.61 & 0.47 / 0.99 & 0.39 / 0.65 & 0.33 / 0.33 & 0.25 / \textbf{0.14} \\ \hline
	LVI\_SAM     & 2.04 / 3.33  & 1.05 / 2.37 & 0.81 / 1.53 & 0.57 / 1.38 & 0.49 / 1.13 & 0.44 / 1.01 \\ \hline
	R2LIVE       & 1.13 / 2.06  & 0.65 / 1.45 & 0.49 / 0.88 & 0.31 / 0.57 & 0.26 / 0.30 & 0.29 / 0.16 \\ \hline
	FAST-LIO2    & 1.31 / 2.22  & 0.69 / 1.34 & 0.49 / 0.89 & 0.31 / 0.59 & \textbf{0.26} / 0.31 & \textbf{0.25} / 0.29 \\ \hline
	VINS-Mono    & 2.44 / 9.35  & 1.31 / 7.28 & 0.99 / 4.63 & 0.66 / 3.68 & 0.56 / 3.24 & 0.48 / 2.73 \\ \hline
%	R3LIVE-LIO   & 1.89 / 2.35  & 1.11 / 1.86 & 0.70 / 1.54 & 0.50 / 1.40 & 0.42 / 1.16 & 0.42 / 1.08 \\ \hline
\end{tabular}
\end{table}

\begin{table}[]
	\centering
	\footnotesize
	\caption{The average time consumption on two different platforms with different configurations. }
	\setlength\tabcolsep{0.5pt}
	\begin{tabular}{|c|c|c|c|c|c|c|c|c|c|c|}
		\hline
		\multicolumn{10}{|c|}{VIO per-frame cost time}                                                                                           & \multirow{3}{*}{\begin{tabular}[c]{@{}c@{}}LIO \\ per-frame\\ cost time\end{tabular}} \\ \cline{1-10}
		Image size           & \multicolumn{3}{c|}{320$\times$256} & \multicolumn{3}{c|}{640$\times$512} & \multicolumn{3}{c|}{1280$\times$1024} &                                                                                       \\ \cline{1-10}
		Pt\_res (m)  & 0.10       & 0.05       & 0.01      & 0.10       & 0.05      & 0.01       & 0.10       & 0.05       & 0.01        &                                                                                       \\ \hline
		On PC (ms) & 7.01       & 10.41      & 33.55     & 11.53      & 14.68     & 39.75      & 13.63      & 17.58      & 43.71       & 18.40                                                                                 \\ \hline
		On OB (ms) & 15.00      & 26.45      & 84.22     & 26.05      & 42.32     & 123.62     & 33.38      & 55.10      & 166.08      & 39.56                                                                                 \\ \hline
	\end{tabular}
	\vspace{-0.8cm}
	\label{tab_rumtime_analysis}
\end{table}

\subsection{Run time analysis}
We investigate the average time consumption among of our system all experiments on two different platforms: the desktop PC (with Intel i7-9700K CPU and 32GB RAM) and a UAV on-board  computer (``OB", with Intel i7-8550u CPU and 8GB
RAM). The detailed statistics are listed in Table. \ref{tab_rumtime_analysis}. The time consumption of our VIO subsystem is affected by two main settings: the image resolution and the point cloud map resolution (``Pt\_res").

\section{Applications with R$^3$LIVE}

\subsection{Mesh reconstruction and texturing}
\begin{figure}[h]
	\centering
	\includegraphics[width=1.0\linewidth]{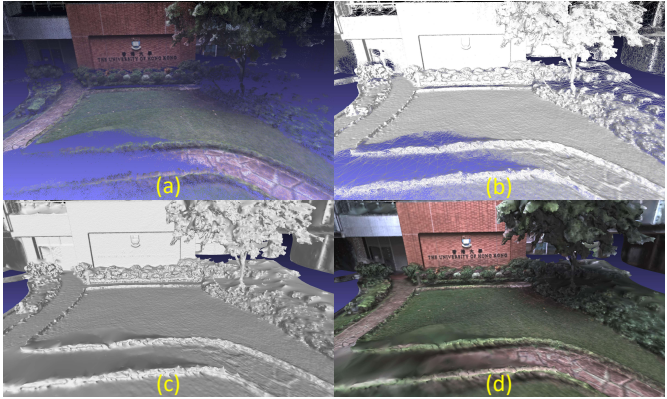}
	\caption{ Figure (a) show the RGB-colored 3D points reconstructed by R$^3$LIVE. Figure (b) and (c) show the wireframe and surface of our reconstructed mesh. Figure (d) show the mesh after texture rendering.}
	\vspace{-0.5cm}
	\label{fig_mesh}
\end{figure}
While R$^3$LIVE reconstructs the colored 3D map in real-time, we also develop software utilities to mesh and texture the reconstructed map offline (see Fig. \ref{fig_mesh}). For meshing, we make use of the Delaunay triangulation and graph cuts \cite{labatut2007efficient} implemented in CGAL\cite{fabri2009cgal}\footnote{\url{https://www.cgal.org/}}. After the mesh construction, we texture the mesh with the vertex colors, with is rendered by our VIO subsystem. 

Our developed utilities also export the colored point map from R$^3$LIVE or the offline meshed map into commonly used file formats such as ``pcd"\footnote{\url{http://pointclouds.org/documentation/tutorials/pcd_file_format.html}}, ``ply"\footnote{\url{https://en.wikipedia.org/wiki/PLY_(file_format)}}, ``obj"\footnote{\url{https://en.wikipedia.org/wiki/Wavefront_.obj_file}} and etc. As a result, the maps reconstructed by R$^3$LIVE can be imported by various of 3D softwares, including but not limited to CloudCompare\cite{girardeau2016cloudcompare}, Meshlab\cite{cignoni2011meshlab}, AutoDesk 3ds Max\footnote{\url{https://www.autodesk.com.hk/products/3ds-max}}, etc.

\subsection{Toward various of 3D applications}
\begin{figure}[h]
	\centering
	\includegraphics[width=1.0\linewidth]{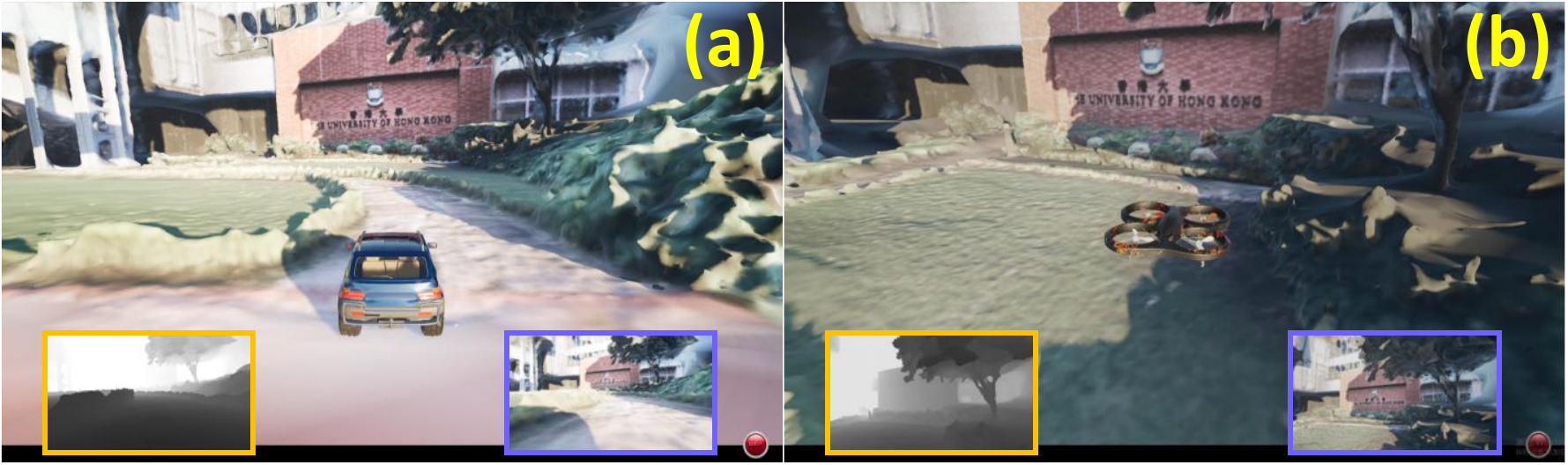}
	\caption{We use the maps reconstructed by R$^3$LIVE to build the car (in (a)) and drone (in (b)) simulator with AirSim. The images in green and blue frameboxes are of the depth, RGB image query from the airsim's API, respectively. }
	\label{fig_airsim}
	\includegraphics[width=1.0\linewidth]{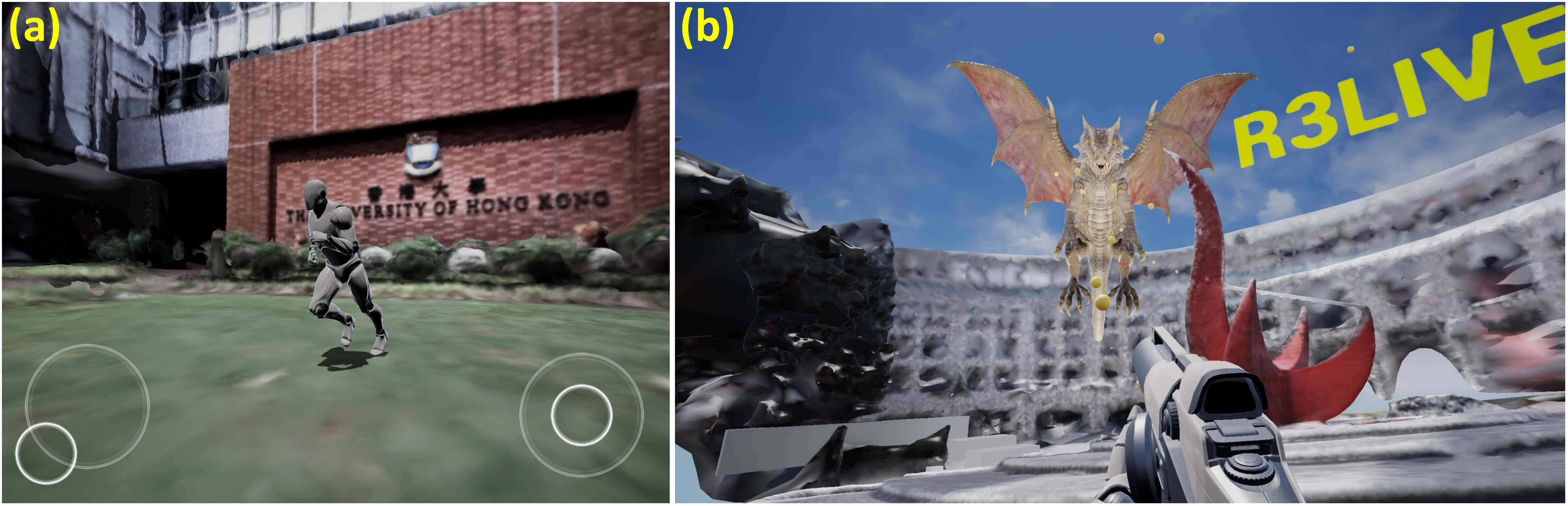}
	\caption{We use the maps built by R$^3$LIVE to develop the video games for mobile platform (see (a)) and desktop PC (see (b)). In (a), the player is controling the actor to explore the campus of HKU. In (b), the player is fighting against the dragon with shoting the rubber balls in the campus of HKUST.}
	\label{fig_ue4_game}
	\vspace{-0.8cm}
\end{figure}

With the developed software utilities, we can export the reconstructed 3D maps to Unreal Engine \footnote{https://www.unrealengine.com} to enable a series of 3D applications. For example, in Fig.\ref{fig_airsim}, we built the car and drone simulator with the AirSim\cite{shah2018airsim}, and in Fig.\ref{fig_ue4_game}, we use our reconstructed maps for developing the video games for desktop PC and mobile platform. For more details about our demos, we refer the readers to watch our video on YoutuBe\footnote{\url{https://youtu.be/4rjrrLgL3nk}}. 

\iffalse
	\section{Acknowledgment}
	The authors would like to thank DJI Co., Ltd\footnote{\url{https://www.dji.com}}. for donating devices and research found.
\fi
%    \clearpage
    \bibliography{ral_2021_r3live}
    \vspace{-0.3cm}

\end{document}